\documentclass[11pt]{article}

\usepackage{geometry}
\usepackage{amsmath}
\usepackage{graphicx}
\usepackage{xurl}
\usepackage{hyperref}
\usepackage{authblk}
\usepackage{setspace}
\usepackage{amssymb}
\usepackage{caption}
\usepackage{subcaption}
\usepackage{booktabs}
\usepackage{longtable}

\usepackage[
backend=biber,
style=alphabetic, 
sorting=ynt
]{biblatex}
\newcommand{\TopologySubsetCount}{10}
\newcommand{\TopologyAuditComparisonCount}{1,592}
\newcommand{\TopologyAuditMaximumAbsoluteDelta}{0}

\newcommand{\ArxivDireAutoKnnGraphSeconds}{12.2}
\newcommand{\ArxivDireAutoProfileSteadySeconds}{13.8}
\newcommand{\ArxivDireIvfFlatProfileSteadySeconds}{44.0}
\newcommand{\ArxivDireAutoInitializationSeconds}{0.16}
\newcommand{\ArxivDireAutoLayoutSeconds}{1.30}
\newcommand{\ArxivDireAutoChunkedFallbackCalls}{0}
\newcommand{\ArxivDireAutoSpeedupVsIvfFlat}{3.19}
\newcommand{\ArxivDireBackendGraphOverlap}{0.840}
\newcommand{\ArxivCumlTsneSteadySeconds}{31.2}

\newcommand{\ArxivCumlUmapSteadySeconds}{39.9}

\newcommand{\ArxivDireAutoSteadySeconds}{13.8}

\newcommand{\TenxCumlTsneSteadySeconds}{33.2}

\newcommand{\TenxCumlUmapSteadySeconds}{14.4}

\newcommand{\TenxDireAutoSteadySeconds}{7.51}

\newcommand{\TenxDireAutoAdjacencyPreserved}{18}
\newcommand{\TenxDireAutoAdjacencyIntroduced}{42}
\newcommand{\TenxDireAutoAdjacencyCellWeighted}{0.328}
\newcommand{\TenxDireSpectralAdjacencyPreserved}{26}

\newcommand{\TenxDireSpectralAdjacencyCellWeighted}{0.688}
\newcommand{\TenxCumlUmapAdjacencyPreserved}{29}
\newcommand{\TenxCumlUmapAdjacencyIntroduced}{31}
\newcommand{\TenxCumlUmapAdjacencyCellWeighted}{0.719}
\newcommand{\TenxDireAutoAdjacencyFilteredEqual}{0.255}
\newcommand{\TenxDireAutoAdjacencyKOne}{0.200}
\newcommand{\TenxDireAutoAdjacencyKTwo}{0.300}
\newcommand{\TenxDireAutoAdjacencyKThree}{0.300}
\newcommand{\TenxCumlUmapAdjacencyFilteredEqual}{0.510}
\newcommand{\TenxCumlUmapAdjacencyKOne}{0.200}
\newcommand{\TenxCumlUmapAdjacencyKTwo}{0.450}
\newcommand{\TenxCumlUmapAdjacencyKThree}{0.483}
\newcommand{\SmallAtlasUmapComparisonCount}{12}
\newcommand{\SmallAtlasDireLowerVsUmapCount}{10}

\newcommand{\SmallAtlasDireIntervalLowerVsUmapCount}{7}
\newcommand{\SmallAtlasDireFivePctLowerVsUmapCount}{7}
\newcommand{\SmallAtlasTsneComparisonCount}{12}
\newcommand{\SmallAtlasDireLowerVsTsneCount}{6}
\newcommand{\SmallAtlasTsneLowerVsDireCount}{6}

\newcommand{\ArxivDireBetaZeroDtwReductionVsUmapPct}{32.8}
\newcommand{\ArxivDireBetaZeroDtwReductionVsTsnePct}{33.0}
\newcommand{\ArxivDireBetaOneDtwReductionVsUmapPct}{71.6}
\newcommand{\ArxivDireBetaOneDtwReductionVsTsnePct}{67.0}

\newcommand{\ArxivTsneMinusDireBetaZeroDtwGapInObservedSd}{5.98}

\newcommand{\ArxivTsneMinusDireBetaOneDtwGapInObservedSd}{9.43}
\newcommand{\ArxivTsneMinusDireBetaZeroQNinetyNineDtwGap}{-1.0407}
\newcommand{\ArxivTsneMinusDireBetaZeroQNinetyNineDtwGapSd}{0.9596}

\newcommand{\ArxivTsneMinusDireBetaOneQNinetyNineDtwGap}{0.0924}
\newcommand{\ArxivTsneMinusDireBetaOneQNinetyNineDtwGapSd}{0.1154}

\newcommand{\ArxivUmapMinusDireBetaZeroDtwGapInObservedSd}{11.02}

\newcommand{\ArxivUmapMinusDireBetaOneDtwGapInObservedSd}{4.10}
\newcommand{\ArxivUmapMinusDireBetaZeroQNinetyNineDtwGap}{0.0113}
\newcommand{\ArxivUmapMinusDireBetaZeroQNinetyNineDtwGapSd}{0.0591}

\newcommand{\ArxivUmapMinusDireBetaOneQNinetyNineDtwGap}{0.0899}
\newcommand{\ArxivUmapMinusDireBetaOneQNinetyNineDtwGapSd}{0.1174}

\title{Dimensionality reduction for homological stability \\ and global structure preservation}

\author[1]{Alexander Kolpakov}
\author[2]{Igor Rivin}  
\affil[1]{University of Austin, Austin TX, USA;
\texttt{akolpakov@uaustin.org}}
\affil[2]{Temple University, Philadelphia PA, USA;  
\texttt{rivin@temple.edu}}

\date{}  

\begin{document}

\maketitle

\begin{abstract}
We propose DiRe, a force--directed dimensionality reduction framework designed to preserve global structure and homological features while remaining practical on modern hardware. The method combines an initial embedding with a graph--based layout optimization and evaluates the resulting low--dimensional representation using local distortion, context preservation, and persistent homology measures. Across the benchmark suite considered here, DiRe provides a complementary tradeoff to UMAP and tSNE: it is designed less as a purely local visualization heuristic and more as a framework for embeddings whose large--scale geometry can be quantified through Betti curves and persistence diagrams.
\end{abstract}

\section{Introduction}
Traditional dimensionality reduction techniques such as UMAP and tSNE \cite{umap,vandermaaten-tsne} are widely used for visualizing high--dimensional data in lower--dimensional spaces, usually 2D and sometimes 3D. Other uses include dimensionality reduction to other, possibly higher and thus non--visual dimensions, for the subsequent use of classifiers such as SVMs.

These methods are designed primarily around local neighbour relations, so
their large--scale organization must be interpreted with care; runtime and
sensitivity also depend on the implementation and configuration
\cite{kobak2019art,pachter}.

Both tSNE and UMAP can change their large--scale organization substantially with initialization \cite{kobak-linderman-initialization}. The original exact tSNE formulation has a quadratic interaction cost, but modern interpolation and FFT--based implementations, including FIt--SNE and openTSNE, remove that formulation as a practical description of current large--scale tSNE software \cite{linderman-fitsne,opentsne}. UMAP is likewise designed around sparse graph optimization and negative sampling rather than explicit evaluation of every non--edge. We therefore do not use ``tSNE is quadratic'' or ``UMAP alone loses global structure'' as premises for DiRe.

DiRe addresses these challenges by combining an initial low--dimensional embedding with a force--directed layout driven by the $k$--nearest--neighbor graph of the original data. The goal is not to replace every use of UMAP or tSNE, but to provide an alternative whose optimization and diagnostics are explicitly tied to preservation of large--scale and homological structure.

DiRe belongs to the broad family of initialized neighbor embeddings. The existence of PCA or spectral initialization is not itself a novelty claim. The object of study is the particular simplified force law, the bounded random--negative approximation used by DiRe, the separation of initialization from refinement, and quantitative evaluation that treats topology as one diagnostic rather than inferring quality from a scatter plot.

The computational results in this manuscript use DiRe--RAPIDS \cite{dire-rapids}. Its GPU stack includes PyTorch \cite{pytorch}, optional PyKeOps acceleration \cite{pykeops}, and RAPIDS/cuVS components \cite{cuml,cuvs}.

The contribution of this paper is therefore twofold. First, it describes the algorithmic framework and the homological stability motivation behind DiRe. Second, it evaluates the framework with metrics that measure local distortion, classification context, and persistent homology of the original data and its embedding.

\subsection{Relation to existing methods}

tSNE, UMAP, and DiRe can all be viewed as neighbor embeddings governed by a balance between attraction and repulsion \cite{vandermaaten-tsne,umap,bohm-attraction-repulsion}. Spectral initialization is the UMAP default, while PCA initialization is common in modern tSNE practice \cite{kobak2019art,kobak-linderman-initialization,opentsne}. DiRe does not claim either initialization as new. Its attraction acts on an unweighted $k$NN graph, and its repulsion uses the explicit kernel below on uniformly sampled point indices. Initialization and refinement are separate implementation stages; the reproduction suite measures initializer--only controls by setting the recorded layout iteration count to zero and compares them with the corresponding refined layouts.

PCA and random projections provide analytical reference points because their distortion can be bounded in terms of variance or Johnson--Lindenstrauss distortion, while spectral methods use graph operators related to the same neighborhood graph \cite{belkin-niyogi-laplacian,coifman-lafon-diffusion}. The topological evaluation is also not presented as having been invented in isolation: persistent homology has previously been used to assess dimensionality--reduction schemes \cite{rieck-leitte-2015}, and rank--based neighborhood criteria have a substantial prior literature \cite{lee-verleysen-quality}. Our contribution is a reproducible application of complementary local, context, centroid, and homological diagnostics to DiRe and its baselines.

\section{Main methods}

Let $X = \{x_1,\ldots,x_n\} \subset \mathbb{R}^D$ be the input point cloud, represented in practice as an $n \times D$ numerical array. DiRe performs the following main steps:

\begin{enumerate}
    \item \textbf{Constructing the neighborhood graph:} Create the kNN graph of $X$, say $\Gamma$, for a given number of neighbors $k$ ($=$ \texttt{n\_neighbors}). The released DiRe--RAPIDS implementation can use PyTorch, PyKeOps, or RAPIDS/cuVS for this step, subject to the selected backend, metric, and available hardware. The selected input metric governs this graph construction; forces in the target coordinates remain Euclidean.
    \item \textbf{Initial dimension reduction:} Produce $Y \subset \mathbb{R}^d$, the initial embedding of $X$, for a positive user--specified \texttt{n\_components}$=d$. The released choices are \texttt{random}, which multiplies $X$ by a Gaussian projection matrix with column--normalised directions; \texttt{spectral}, which uses nontrivial eigenvectors of the normalised adjacency of the symmetrised unweighted kNN graph; and \texttt{pca}, whose implementation is backend dependent. In the cuVS backend used for the H100 profile, the \texttt{pca} pathway dispatches to cuML PCA for at most 100 input features and cuML TruncatedSVD above 100 input features. Each initial coordinate is then centred and standardised.
    \item \textbf{Layout optimization:} Adjust the lower--dimensional embedding $Y$ to conform to the similarity structure of the higher--dimensional data $X$. This is done by using a force layout in which the role of ``forces'' is played by probability kernels; their scale is controlled by parameters such as \texttt{min\_dist} and \texttt{spread}.
\end{enumerate}

In DiRe--RAPIDS, users can instantiate PyTorch, memory--efficient PyTorch/PyKeOps, or cuVS backends directly; alternatively, \texttt{create\_dire} selects an implementation according to available dependencies and hardware. The initialization and optimization stages are separate in the implementation. The reproduction suite obtains an initializer--only output with \texttt{max\_iter\_layout=0} and a refined output with a positive iteration budget, so the layout step can be studied without claiming that the estimator returns both arrays from one call.

The initial embedding can be one of the following three choices, each of which we discuss in more detail.

\subsection{Random Projection embedding} This embedding is based on the Johnson--Lindenstrauss lemma \cite{JL}.

\medskip

\textbf{Johnson--Lindenstrauss Lemma (Probabilistic Form)}

Given \(0 < \epsilon < 1\) and an integer \(n\), let \(X\) be a set of \(n\) points in \(\mathbb{R}^d\). For a random linear map \(f: \mathbb{R}^d \to \mathbb{R}^k\) where \(k = O\left(\frac{\log n}{\epsilon^2}\right)\), with high probability, for all \(u, v \in X\),

\[
(1 - \epsilon) \|u - v\|^2 \leq \|f(u) - f(v)\|^2 \leq (1 + \epsilon) \|u - v\|^2.
\]

The value \[\text{dist}(f) = \frac{\|f(u) - f(v)\|}{\|u - v\|}\] is called the \textit{distortion} of $f$, and is expected to be close to $1.0$ for a good quality embedding. 

The Johnson--Lindenstrauss lemma says that a random projection from $\mathbb{R}^d$ to $\mathbb{R}^k$ has, with high probability, distortion close to $1$ when $k$ is sufficiently large relative to $\log n$. This guarantee does not imply small distortion when $k$ is restricted to the visualization range ($k=1, 2, 3$). The method is nevertheless simple and computationally inexpensive.

The image of a projection $p: \mathbb{R}^d \rightarrow \mathbb{R}^k$ may be cluttered when $k \ll d$. Indeed, the total variance of $Y = p(X) \in \mathbb{R}^{n\times k}$ may be much lower than the total variance of $X \in \mathbb{R}^{n\times d}$ unless the projection captures the dominant variance directions.

\subsection{Principal Component Analysis embedding}

The discussion above motivates projections that preserve as much variance as possible. The classical method for this purpose is Principal Component Analysis.

Assume that $X$ is centered, so that each column of $X$ has zero mean. Let its covariance matrix be $\text{Cov}(X) = \frac{1}{n-1} X^t X$. PCA is computed from the singular value decomposition
\[ X = U \, \Sigma \, W^t. \]
For a target relative residual $0<\varepsilon<1$, choose $k\ll d$ so
that the omitted singular values satisfy
\[
  \sum_{i>k}\sigma_i^2\leq\varepsilon^2\|X\|_F^2,
  \qquad\text{equivalently}\qquad
  \sum_{i=1}^k\sigma_i^2\geq(1-\varepsilon^2)\|X\|_F^2.
\]

Let $\Sigma_k$ be the rank $k$ truncation of $\Sigma$ preserving the dominant singular values above, and let $U_k$ and $W_k$ be the respective truncations of $U$ and $W$. Then the rank $k$ approximation of $X$ is
\[\widehat{X} =  U_k\, \Sigma_k\, W^t_k \]
and the PCA embedding of $X$ is
\[ X_k = X\, W_k. \]

If the dominant topological features of $X$ are represented in the leading
$k$ principal directions, then the persistence diagrams of $X$ can be
compared with those of the projected point cloud $X_k$. This comparison is
exact when $X$ lies in an affine $k$--dimensional subspace of
$\mathbb{R}^d$. More generally, $\widehat X$ lies in the leading
$k$--dimensional subspace and is isometric to $X_k=\widehat XW_k=XW_k$.
Vietoris--Rips stability therefore gives, for each homological dimension $q$
\cite{chazal-persistence},
\[
\begin{aligned}
d_b\!\left(D_q(X),D_q(X_k)\right)
&=d_b\!\left(D_q(X),D_q(\widehat X)\right)\\
&\leq 2d_H(X,\widehat X)
\leq 2\max_i\|x_i-\widehat x_i\|_2\\
&\leq 2\|\widehat X-X\|_F
\leq2\varepsilon\|X\|_F .
\end{aligned}
\]
The factor two is the standard Vietoris--Rips stability constant; it should
not be confused with an assertion that PCA preserves arbitrary topology
exactly.

In this sense, PCA gives a useful reference case: the homological distortion is controlled by the low--rank approximation error when the retained coordinates capture the relevant geometry.

\subsection{Spectral Laplacian embedding}

Spectral Laplacian embeddings are commonly used for manifold learning from local graph structure \cite{tenenbaum-isomap,belkin-niyogi-laplacian}. Diffusion maps use related graph operators to organize data by diffusion geometry \cite{coifman-lafon-diffusion}. Since the $kNN$--graph $\Gamma$ of $X$ is already available in DiRe, a Laplacian embedding can be computed with standard graph methods.

\subsection{Force--directed layout}

After the initial embedding, DiRe applies an iterative force--directed layout in
order to correlate the local structure of the embedding $Y$ with the local
structure of the high--dimensional dataset $X$. Its coefficients use a
heavy--tailed low--dimensional kernel related to the kernels used by UMAP and
tSNE; the kernel family itself is not claimed as novel.

Namely, a simple density function 
\[\varphi(x) = \frac{1}{1 + a\, \|x\|^{2b}}\]
is fit to \texttt{min\_dist} $=\delta$ and \texttt{spread} $=\sigma$ parameters, so that 
\[\varphi(x) \approx 1.0\]
within $\|x\| < \delta$, and then decays exponentially as
\[\varphi(x) \approx \exp( - (\|x\| - \delta)/\sigma )\]
outside of the $\delta$--ball. 

For $r_{ij}=\|y_j-y_i\|$ and unit direction
$u_{ij}=(y_j-y_i)/(r_{ij}+\varepsilon)$, the released force kernel uses the
attractive coefficient
\[
A(r_{ij})=\varphi(1/r_{ij})
          =\frac{r_{ij}^{2b}}{r_{ij}^{2b}+a}
\]
on the directed $kNN$ edges. Its repulsive coefficient is
\[
R(r_{ij})=\varphi(r_{ij})\exp(-r_{ij}/c)
          =\frac{\exp(-r_{ij}/c)}{1+a r_{ij}^{2b}},
\]
where $c$ is the implementation's \texttt{cutoff} parameter. Thus one
stochastic force update has the form
\[
F_i=\sum_{j\in N_X(i)} A(r_{ij})u_{ij}
    -\sum_{j\in S_i} R(r_{ij})u_{ij},
\]
followed by componentwise force clipping and a linearly decreasing step size.
The final coordinates are recentered and standardized by output coordinate.

The benchmarked DiRe--cuVS implementation does \emph{not} enumerate all
non--edge pairs. For each point and iteration it draws
\[
q = \min\{r k,n-1\}
\]
indices uniformly with replacement from the point-index range, where
$r=\texttt{neg\_ratio}$ (default $8$), and evaluates repulsion only for that
bounded sample $S_i$. The random seed is recorded in every benchmark result.

The layout is run for a prescribed iteration budget. Parameters such as \texttt{n\_neighbors}, \texttt{min\_dist}, \texttt{spread}, backend choice, random seed, and floating--point precision affect the final embedding and are part of any reproducible specification. In applications these parameters can be tuned against a chosen validation criterion, for example by Bayesian optimization packages such as Hyperopt~\cite{hyperopt} or Optuna~\cite{optuna}. In the benchmarks below we instead use fixed, reported protocol choices so that the comparison is reproducible and not dataset--specific hyperparameter search.

\subsection{Computational scaling}

Let $n$ be the number of points, $d$ the output dimension, $k$ the neighbor count, $r$ the negative--sample ratio, and $T$ the layout iteration count. Once the neighbor graph and initialization have been constructed, one DiRe layout iteration evaluates $nk$ attractive and at most $nrk$ repulsive interactions. Its layout work is therefore
\[
O\!\left(T n k(1+r)d\right),
\]
and the graph plus sampled force tensors require memory linear in $n$ when $k$, $r$, and $d$ are fixed. This is a sampled sparse layout, not a hidden all--pairs computation.

Neighbor--graph construction is a separate stage whose cost depends on the
selected search engine and data dimension. The primary large--data DiRe
experiment uses the package's production \texttt{cuvs\_index\_type=auto}
policy and records the effective cuVS index selected at every size. The
previously measured forced IVF--Flat configuration is retained as a separately
named control rather than overwritten. A fresh paired backend audit records
synchronized graph, initialization, and layout times for both policies,
together with fixed--query neighbor--set overlap and force--fallback status.
Thus, the linear bound above describes the layout stage; we do not mislabel
approximate neighbor search as a universal worst--case linear kNN algorithm.
The scaling experiments report end--to--end \texttt{fit\_transform} time and
peak incremental device--memory use sampled through NVML, so graph
construction, initialization, compilation, data transfer, and layout are all
included in the measured result.
    
\section{Quantitative measures}

We use several quantitative measures to assess the quality of DiRe embeddings, and to compare the algorithmic framework to other embedding techniques available, such as tSNE in its cuML implementation, as well as UMAP in both its original and cuML implementations.

\subsection{Global--structure diagnostics}

We use several types of persistent--homology measures, following earlier work on topological evaluation of dimensionality reduction \cite{rieck-leitte-2015}. One type is related to \textit{persistence diagrams}, and the other is related to the \textit{Betti curves} derived from them. These representations are useful in different ways, but no topological score is treated as a universal embedding-quality criterion.

\subsubsection{Persistence diagrams}

Persistence diagrams are classical representations of the ``birth--death'' pairs in persistent homology. The reader may find more information in \cite{ZC, zomorodian2005topology}, while we give a brief description below.

The \textit{Vietoris--Rips complex} at scale \(t \geq 0\), denoted
\(\operatorname{VR}_t(X)\), is the simplicial complex defined as
\[
\operatorname{VR}_t(X) = \left\{ \sigma \subseteq X \mid \|x_i - x_j\| \leq t \text{ for all } x_i, x_j \in \sigma \right\},
\]
which means a simplex \(\sigma\) is contained in \(\operatorname{VR}_t(X)\) if all its vertices are pairwise within distance~\(t\).

Consider a non--decreasing sequence of scales \(\{t_i\}_{i \in I}\), where \(I\) is an index set (often \(I = \mathbb{R}_{\geq 0}\)), generating a \textit{filtration} of Rips complexes:

\[
\operatorname{VR}_{t_0}(X) \subseteq \operatorname{VR}_{t_1}(X) \subseteq \operatorname{VR}_{t_2}(X) \subseteq \cdots.
\]

For each \(k \geq 0\) and scale \(t_i\), compute the \(k\)-th homology group \(H_k\left( \operatorname{VR}_{t_i}(X); \mathbb{F} \right)\) with coefficients in a field \(\mathbb{F}\) (commonly \(\mathbb{F} = \mathbb{Z}/2\mathbb{Z}\), or another finite field). The inclusion maps induce homomorphisms between homology groups:

\[
f_{s,t}^k: H_k\left( \operatorname{VR}_{s}(X); \mathbb{F} \right) \rightarrow H_k\left( \operatorname{VR}_{t}(X); \mathbb{F} \right), \quad \text{for } s \leq t.
\]

A persistent \(k\)-dimensional homology class \(\alpha\) is an element of \(H_k\left( \operatorname{VR}_{s}(X); \mathbb{F} \right)\) that persists across multiple scales. The \textit{birth time} \(b_\alpha\) and \textit{death time} \(d_\alpha\) of \(\alpha\) are defined as

\begin{itemize}
    \item \(b_\alpha\): the smallest \(t\) where \(\alpha \neq 0\) appears in \(H_k\left( \operatorname{VR}_{t}(X); \mathbb{F} \right)\);

    \item \(d_\alpha\): the smallest \(t > b_\alpha\) where \(f_{b_\alpha, t}^k(\alpha) = 0\).
\end{itemize}

For homology classes that persist indefinitely, we set their death time to infinity: \(d_\alpha = \infty\).

The \textit{persistence diagram} \(D_k\) is the multiset of points \((b_\alpha, d_\alpha)\) in the extended plane \(\mathbb{R}^2 \cup \{\infty\}\):

\[
D_k(X) = \left\{ (b_\alpha, d_\alpha) \in (\mathbb{R} \times (\mathbb{R} \cup \{\infty\})) \mid \alpha \in H_k(X) \text{ with birth } b_\alpha \text{ and death } d_\alpha \right\},
\]
where we use $H_k(X)$ to denote the $k$--dimensional persistent homology group defined above.

Features with longer lifespans \(d_\alpha - b_\alpha\) are considered topologically significant. In particular, features with \(d_\alpha = \infty\) represent persistent homological features that never disappear within the considered scale range, indicating essential topological structures of the space.

The usual metrics between two diagrams $D = D_k(X)$ and $D' = D_k(Y)$
include the bottleneck and Wasserstein distances, cf.\
\cite{persim-docs}. Because an arbitrary output scale is inherent to
visualization algorithms, an unnormalised diagram distance can primarily
measure dilation rather than homology preservation. We therefore use the
diameter--normalised bottleneck distance developed specifically for
dimension--reduction comparisons \cite{may-normalized-bottleneck-2024}:
\[
d_N^{(k)}(X,Y)
=
d_b\!\left(
D_k\!\left(X/\operatorname{diam}(X)\right),
D_k\!\left(Y/\operatorname{diam}(Y)\right)
\right).
\]
The single essential $H_0$ class is common to both diagrams and is excluded
from the finite--bar matching.
For the remaining $H_0$ bars, all births are zero. If their sorted finite death
times are $a_1\leq\cdots\leq a_m$ and
$b_1\leq\cdots\leq b_m$, we compute the bottleneck distance exactly as
\[
\min_{0\leq u\leq m}
\max\!\left\{
  \frac{1}{2}\max_{1\leq i\leq u}\{a_i,b_i\},
  \max_{u<i\leq m}|a_i-b_i|
\right\},
\]
where either maximum over an empty set is zero. Here $u$ is the number of
shortest bars in each diagram matched to the diagonal; the remaining sorted
tails are matched in order. This scalar specialization avoids constructing a
general diagram-matching problem for roughly 4,000 zero-birth bars. The
$H_1$ bottleneck distance uses the compiled exact GUDHI matcher.

\subsubsection{Betti curves}

Betti curves mark the progression of the persistence Betti numbers with the filtration level increase. This representation was first studied in detail in \cite{Giusti-et-al}. Let $X$ be a point cloud and $\mathrm{Rips}_t(X)$ be its Rips complex of level $t \geq 0$. Then the $k$--th Betti curve is
\[
\beta_k(t) = \mathrm{rank}\, H_k(\mathrm{VR}_t(X)).
\]

Several metrics can be used to compare two Betti curves. We retain the
historical FastDTW approximation to Dynamic Time Warp distance
\cite{salvador-fastdtw,fastdtw-docs}, normalised by the number of sampled
points. This provides continuity with the archived experiments, while the
normalised bottleneck distance above supplies a direct persistence--diagram
comparison with an established scale--invariant definition.

\subsubsection{Reported persistence diagnostics}

For the benchmark outputs configured with target dimension $d=2$, let
$X \in \mathbb{R}^{n\times D}$ be the input and
$Y=f(X)\in\mathbb{R}^{n\times 2}$ the recorded output. For $k=0,1$, we report
both $d_N^{(k)}(X,Y)$ and the DTW distance between the corresponding Betti
curves. Lower values indicate closer homological summaries; neither quantity
is interpreted as a complete measure of global structure.

For the large--data protocol, $N$ denotes the full dataset size and $m$ the
number of row indices in one topology subset. The primary computation uses
\TopologySubsetCount{} independently seeded, fixed stratified subsets with
$m=4000$. Each subset is applied identically to the high--dimensional input and
every layout; these are paired diagnostic subsets, not reducer fits or
biological specimens. DiRe and every newly executed baseline are fitted
to every observation. The separately labelled Cell Ranger tSNE reference was
released by 10x with coordinates for all cells; it is not presented as a
reducer rerun on our common 20--feature input and is excluded from runtime
comparisons. Fixed row-index subsetting is confined to the persistence
diagnostic: the
1,306,127--cell input alone has
approximately $8.53\times 10^{11}$ unordered point pairs before any
higher--dimensional Vietoris--Rips simplices are considered. The subset
coordinates, row indices, persistence diagrams, and individual repeated scores
are included in the checksummed result bundle.

To test whether the former 2,000-row diagnostic produced a subset-size
artifact, we additionally evaluate each fixed full-data layout on nested,
stratified row--index subsets with $m\in\{1000,2000,4000\}$ drawn within the
first primary subset. The layout and all method parameters are held fixed in
this calculation. It is reported as subset-size sensitivity rather than pooled
with the \TopologySubsetCount{} independent primary subsets.

We report row--subset and layout--seed variability separately. The canonical
seed-42 layout is evaluated on all \TopologySubsetCount{} fixed row--index
subsets. Independently, three full layouts (seeds 42, 43, and 44) are evaluated
on the first fixed subset for each stochastic nonlinear GPU method. Thus an
error bar over row--index subsets is not presented as though it measured
optimizer stochasticity, or vice versa.

Diameter normalisation has a precise scale--invariance interpretation, but it
can be sensitive to one extreme point
\cite{may-normalized-bottleneck-2024}. We therefore also report a
predeclared sensitivity calculation in which each point cloud is divided by
the 99th percentile of its pairwise distances. This Q99 calculation has no
claim to the pseudometric guarantees of $d_N$; it is explicitly a robustness
diagnostic. The historical unit--diameter, FastDTW--approximated Betti--curve
distances are retained, not
silently replaced after observing the results.

\subsection{Measuring the local structure}

We use two local measures: embedding stress and neighborhood preservation ratio. 

\subsubsection{Embedding stress}

Let $X \in \mathbb{R}^{n\times d}$ be a set of $n$ vectors in $\mathbb{R}^d$ with $kNN$--graph $\Gamma$. Let $Y = f(X) \in \mathbb{R}^{n\times k}$ be its embedding. For each edge $e = (x,y)$ in $\Gamma$ we have its local embedding stress measured by 
\[
\lambda(e) = \left| 1 - \frac{\|x-y\|_2}{\|f(x) - f(y)\|_2}  \right|
\]

Let $\Lambda = \Lambda(f, \Gamma) = (\lambda(e))_{e \in \text{edges}(\Gamma)}$ be the sequence of local edge stresses.  The total embedding stress of $\Gamma$ under $f$ is then computed as 
\[
\sigma(f, \Gamma) = \frac{\sqrt{\mathrm{Var}[ \Lambda]}}{\mathbb{E}[\Lambda]}.
\]

Note that $\sigma$ is scale--invariant in the sense that if all edge lengths of $\Gamma$ are being rescaled under $f$, then  $\sigma(f, \Gamma) = 0$, as there is no \textit{relative} change of the internal metric of $\Gamma$.

\subsubsection{Neighborhood preservation}

Given $X$ and its $kNN$ graph $\Gamma_X := \Gamma$, let us consider $Y = f(X)$ and the $kNN$ graph $\Gamma_Y$ of $Y$. The neighborhood preservation measure counts how many neighbours of a vertex $x$ in $\Gamma_X$ are mapped to neighbours of $f(x)$ in $\Gamma_Y$. That is, let $N_X(x)$ be the neighbor indices for a vertex $x$ of $\Gamma_X$. Let the vertex indices in $Y$ be inherited from $X$, that is  $y_i = f(x_i)$, for $x_i$ a vertex in $\Gamma_X$. Let $N_Y(f(x))$ be the neighbor indices for the embedded point $f(x)$ in $\Gamma_Y$.

Then the neighborhood preservation sequence of $f$ is given by
\[
N(X, f) = \left( \frac{\# \bigl(N_X(x)\cap N_Y(f(x))\bigr)}{\# N_X(x)} \right)_{x \in X}.
\]
The neighborhood preservation index is then the pair 
\[
\nu(X, f) = \left( \mathbb{E}[N(X,f)], \sqrt{\mathrm{Var}[N(X,f)]} \right). 
\]
We intentionally compute both the mean and standard deviation in order to see not only how well neighborhoods are preserved, but also how variable the loss of local structure is. 

In our numerical experiments, this quantity is interpreted together with stress and persistence metrics, since local nearest--neighbour agreement and global topological agreement need not rank the methods in the same way.

\subsection{Measuring context loss}

If $X$ is a labeled set of data points, then its \textit{context} is the mutual position, both geometric and topological, of its labeled subsets. A useful lower--dimensional embedding should preserve enough of this context to remain informative about the high--dimensional relationships in $X$.

\subsubsection{Linear SVM classifier accuracy}

If $X$ is labeled, we compare the accuracy of a linear Support Vector Machine trained and tested on $X$ with the corresponding accuracy for $Y$. This provides a simple measure of whether the embedding changes the linear separability of labeled subsets.

Thus, we expect the accuracy $\alpha_X$ of an SVM classifier trained and tested on $X$ to be close to that of one trained and tested on $Y$, denoted $\alpha_Y$. The SVM context loss is defined by
\[
\kappa_{SVM} = \log \min\left\{ \frac{\alpha_X}{\alpha_Y}, \frac{\alpha_Y}{\alpha_X} \right\}
\]

A substantial change in accuracy indicates that the embedding has changed the linear separability of labeled subsets.

\subsubsection{\texorpdfstring{$kNN$}{kNN} classifier accuracy}

We also compare the accuracy of a $kNN$ classifier trained and tested on $X$ to that of one trained and tested on $Y$. This measures whether local class neighborhoods are preserved by the embedding.

Let $\alpha_X$ be the accuracy of a $kNN$ classifier trained and tested on $X$, and let $\alpha_Y$ be the accuracy of a $kNN$ classifier trained and tested on $Y$. The $kNN$ context loss is then defined as 
\[
\kappa_{kNN} = \log \frac{\alpha_Y}{\alpha_X}.
\]

\section{General workflow}
DiRe follows a streamlined workflow that is designed for ease of use and integration into existing data analysis pipelines.

\begin{enumerate}
    \item \textbf{Data Preprocessing:} Scientific preprocessing remains dataset dependent: transformations such as $\log$ or $\sinh^{-1}$, feature selection, and treatment of metadata must follow the data and question. Separately, the released DiRe--RAPIDS estimator has \texttt{normalize=True} by default; this mean--centres the supplied array and divides it by one global maximum--absolute--value scalar for numerical safety, which preserves Euclidean neighbor rankings. The H100 profile sets \texttt{normalize=False} because it supplies already prepared common inputs to every reducer. The benchmark suite records both the preparation and estimator setting; for single--cell datasets, metadata columns are treated as labels or covariates rather than expression features unless stated otherwise.
    
    \item \textbf{Data embedding:} Performed according to the steps described in the previous section (cf.~Methods).  
    
    \item \textbf{Integration and Export:} DiRe--RAPIDS is designed to be integrated into larger machine learning workflows. It exposes a \texttt{fit\_transform} interface, backend selection utilities, memory--efficient variants, and metric utilities for comparing embeddings. 

    \item \textbf{Embedding metrics:} The metrics described above make the final embedding quantitatively comparable across methods and parameter choices, and can also be used as objectives for hyperparameter optimization.
\end{enumerate}

\section{Benchmarking}

A collection of benchmarks is available in the DiRe--RAPIDS repository
\cite{dire-rapids}, some of which are given below. We concentrate on the main
differences between DiRe, tSNE, and UMAP. For the visual comparisons and their
associated metrics, DiRe, tSNE, UMAP, and PCA are instantiated with target
dimension $d=2$ under a common protocol; this is the configuration of those
experiments, not the scope of DiRe. The GPU protocol uses DiRe--RAPIDS and cuML
implementations of tSNE, UMAP, and PCA on the same accelerator \cite{cuml}. A separate CPU--reference panel
reports openTSNE \cite{opentsne} and the original UMAP implementation
\cite{umap-git,umap}. CPU and GPU runtimes are never mixed into a
hardware--normalized ranking.

\subsection{Benchmark protocol}

All quantitative benchmark figures in this manuscript are regenerated from machine--readable run logs. Scatter plots are regenerated from stored coordinates when present; the committed archive also retains canonical embedding PNGs as an explicit fallback. For each dataset and method we record the dataset source, number of points, ambient dimension, preprocessing, random seed, backend, wall--clock \texttt{fit\_transform} runtime, and the local, context, and persistence metrics used below. The scatter plots use a single canonical seed, while repeated metric plots report means and sample standard deviations. In the fresh six--dataset H100 profile, each GPU configuration has 20 seeded fits and each CPU reference has 10. The first successful fit is retained as the configuration cold start; runtime means and sample standard deviations use the remaining 19 GPU or 9 CPU fits rather than combining compilation/JIT startup with steady execution.

The small--suite persistence audit calls the DiRe--RAPIDS GPU rank--based
local--$k$NN atlas evaluator directly on a fixed $m=1000$ row--index subset
for every fit. Each record is required to identify
\texttt{backend="atlas"}, the direct atlas implementation, and
\texttt{prefer\_ripser=False}; this protocol does not fall through to the
wrapper's Ripser preference. Subset seeds are independent of layout seeds, and
the identical indices and their SHA--256 digest are recorded for every method;
consequently the reported comparator-minus-DiRe effects are paired. Means,
sample standard deviations, minima, and every paired gap are retained. The
default DiRe configuration is the fixed comparator.

The million--point profile uses a distinct, explicitly labelled exact persistent--homology diagnostic: \TopologySubsetCount{} paired, fixed--seed row--index subsets of the high--dimensional input and embedding, each with $m=4000$ indices applied identically to the input and every embedding. Three independently seeded full layouts are evaluated separately on the first fixed subset. Runtime comparisons record the GPU model, driver, package versions, cold start, synchronized steady--state fits, and peak incremental device--memory use sampled through NVML. Each large dataset/method/size configuration runs in an isolated process so that allocator state or a failure in one reducer cannot contaminate another.

\begin{figure}[tbp]
    \centering
    \includegraphics[width=\textwidth]{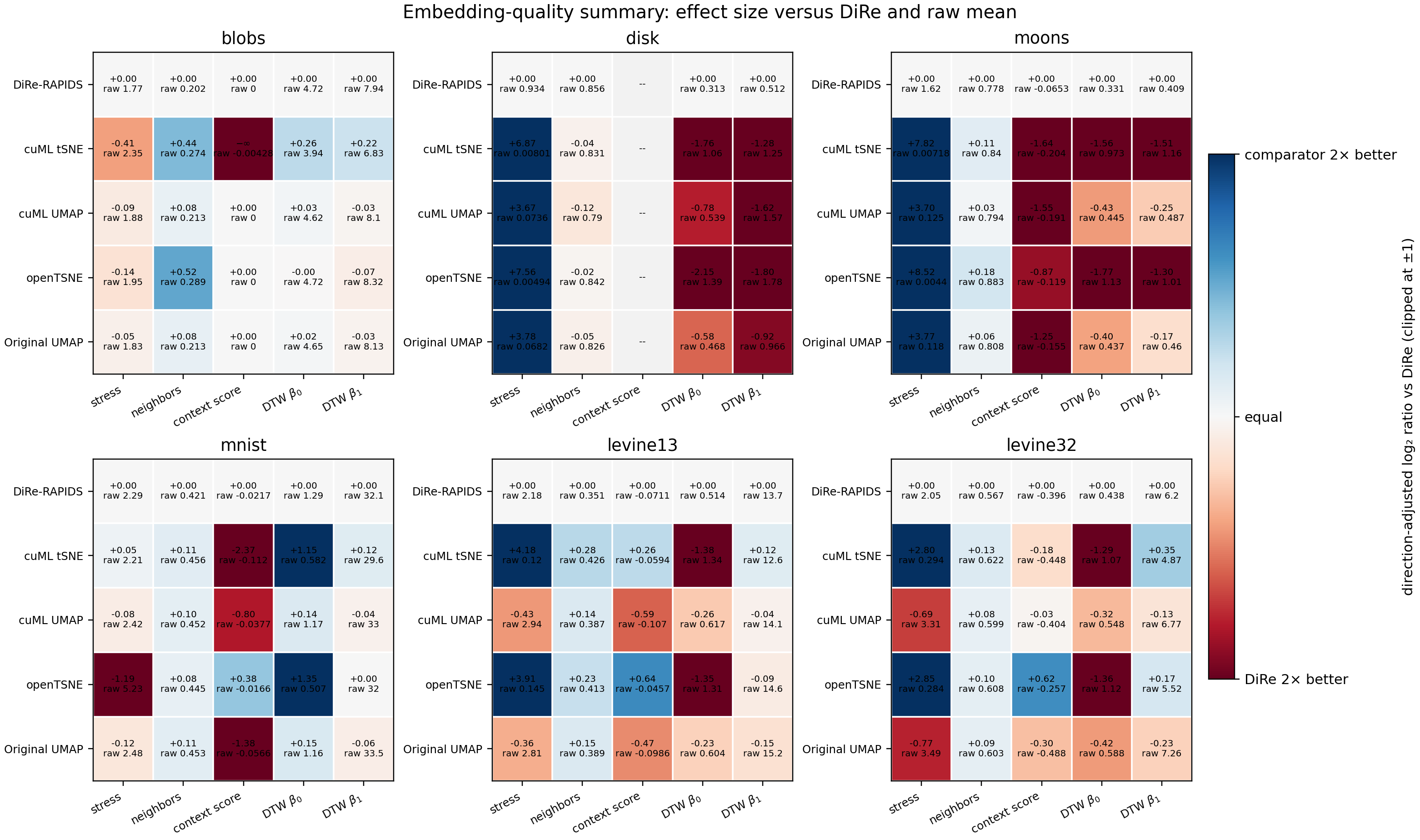}
    \caption{Condensed quality summary for the fresh six--dataset H100
    benchmark. Each
    cell shows the exact raw repeated--run mean and a direction--adjusted
    $\log_2$ ratio against default DiRe. Positive values favour the comparator
    and negative values favour default DiRe; $+1$ and $-1$ denote twofold differences in
    the favourable direction. Lower stress, smaller absolute context-score
    magnitude, and lower Betti--curve discrepancy are better, whereas higher
    neighborhood overlap is better. The printed context means retain the
    signed score defined in the Methods; their absolute magnitudes determine
    the colour and effect. The continuous colour scale is centred at equality
    and clipped at $\pm1$ for legibility; the printed effect values are not
    clipped.}
    \label{fig:archived-quality-summary}
\end{figure}

\begin{figure}[tbp]
    \centering
    \includegraphics[width=\textwidth]{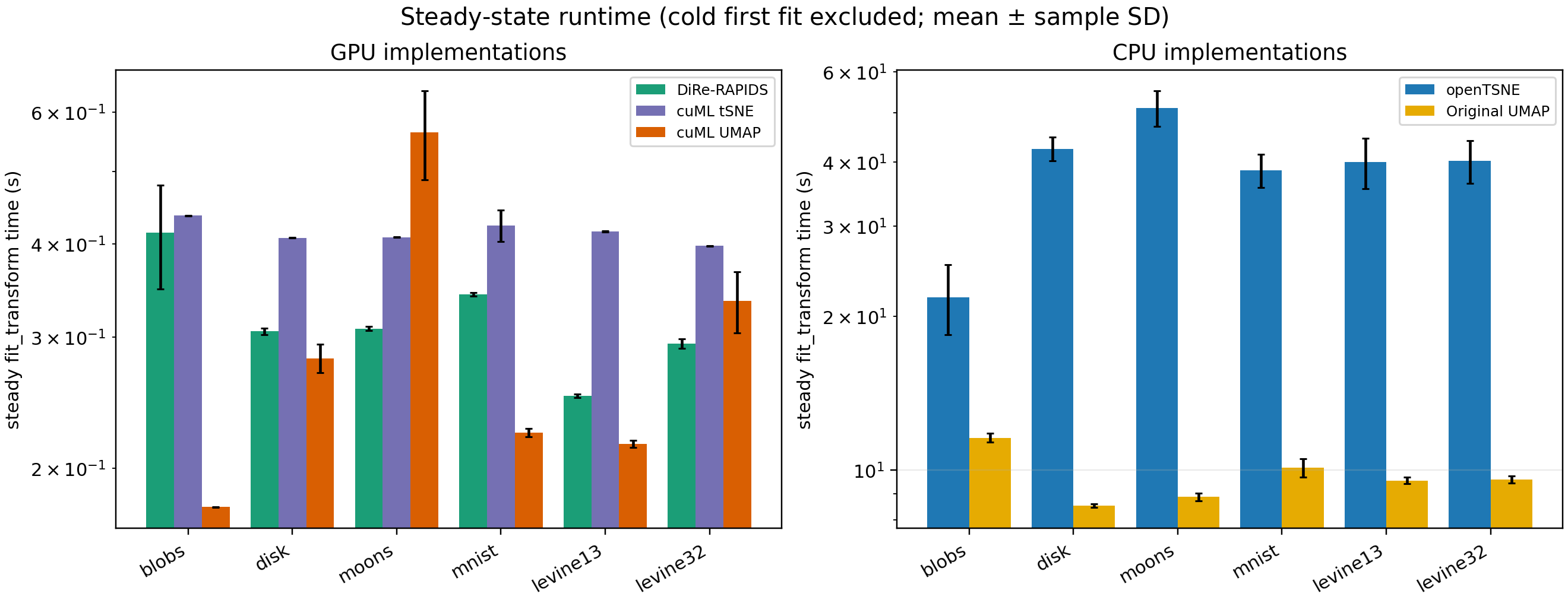}
    \caption{Fresh runtime summary with GPU and CPU implementations in
    separate panels. The GPU panel compares default DiRe--RAPIDS,
    cuML tSNE, and cuML UMAP on the same
    NVIDIA H100 PCIe. The CPU panel is a practical reference and is
    not used for hardware--normalized speed claims. Points and whiskers are
    steady--fit means and sample standard deviations over runs 2--20 for GPU
    configurations and 2--10 for CPU references; the separately retained
    first configuration run is excluded because it can include CUDA
    compilation or CPU JIT startup.}
    \label{fig:archived-runtime-summary}
\end{figure}

\begin{table}[tbp]
  \centering
  \scriptsize
  \resizebox{\textwidth}{!}{%
    \begin{tabular}{@{}llrrr@{}}
\toprule
Dataset & Metric & DiRe & cuML UMAP & cuML t-SNE \\
\midrule
Blobs & $\beta_0$ & 4.720$\pm$0.044 [4.644] & 4.623$\pm$0.035 [4.582] & 3.944$\pm$0.040 [3.882] \\
Blobs & $\beta_1$ & 7.941$\pm$0.382 [7.350] & 8.096$\pm$0.400 [7.250] & 6.831$\pm$0.306 [6.300] \\
\addlinespace
Disk & $\beta_0$ & 0.313$\pm$0.015 [0.280] & 0.539$\pm$0.016 [0.503] & 1.059$\pm$0.019 [1.028] \\
Disk & $\beta_1$ & 0.512$\pm$0.111 [0.389] & 1.571$\pm$0.166 [1.101] & 1.247$\pm$0.227 [0.983] \\
\addlinespace
Moons & $\beta_0$ & 0.331$\pm$0.020 [0.299] & 0.445$\pm$0.037 [0.380] & 0.973$\pm$0.051 [0.894] \\
Moons & $\beta_1$ & 0.409$\pm$0.136 [0.124] & 0.487$\pm$0.203 [0.187] & 1.162$\pm$0.226 [0.841] \\
\addlinespace
MNIST & $\beta_0$ & 1.293$\pm$0.067 [1.202] & 1.170$\pm$0.048 [1.085] & 0.582$\pm$0.036 [0.530] \\
MNIST & $\beta_1$ & 32.078$\pm$3.560 [24.791] & 33.044$\pm$3.341 [26.512] & 29.562$\pm$3.308 [23.192] \\
\addlinespace
Levine13 & $\beta_0$ & 0.514$\pm$0.040 [0.449] & 0.617$\pm$0.049 [0.532] & 1.340$\pm$0.046 [1.272] \\
Levine13 & $\beta_1$ & 13.726$\pm$2.141 [10.242] & 14.132$\pm$2.577 [10.394] & 12.637$\pm$2.435 [7.527] \\
\addlinespace
Levine32 & $\beta_0$ & 0.438$\pm$0.049 [0.354] & 0.548$\pm$0.050 [0.458] & 1.073$\pm$0.055 [0.987] \\
Levine32 & $\beta_1$ & 6.204$\pm$1.042 [4.207] & 6.772$\pm$1.132 [5.031] & 4.873$\pm$1.061 [3.059] \\
\bottomrule
\end{tabular}
}
  \caption{Repeated small--suite atlas--topology discrepancies. Entries are
  mean $\pm$ sample standard deviation [minimum] over 20 GPU fits; lower is
  better. Every method uses the same independently seeded $m=1000$
  row--index subset for the corresponding layout seed. Default DiRe is the
  fixed comparator.}
  \label{tab:small-atlas-topology}
\end{table}

\begin{figure}[tbp]
  \centering
  \includegraphics[width=\textwidth]
    {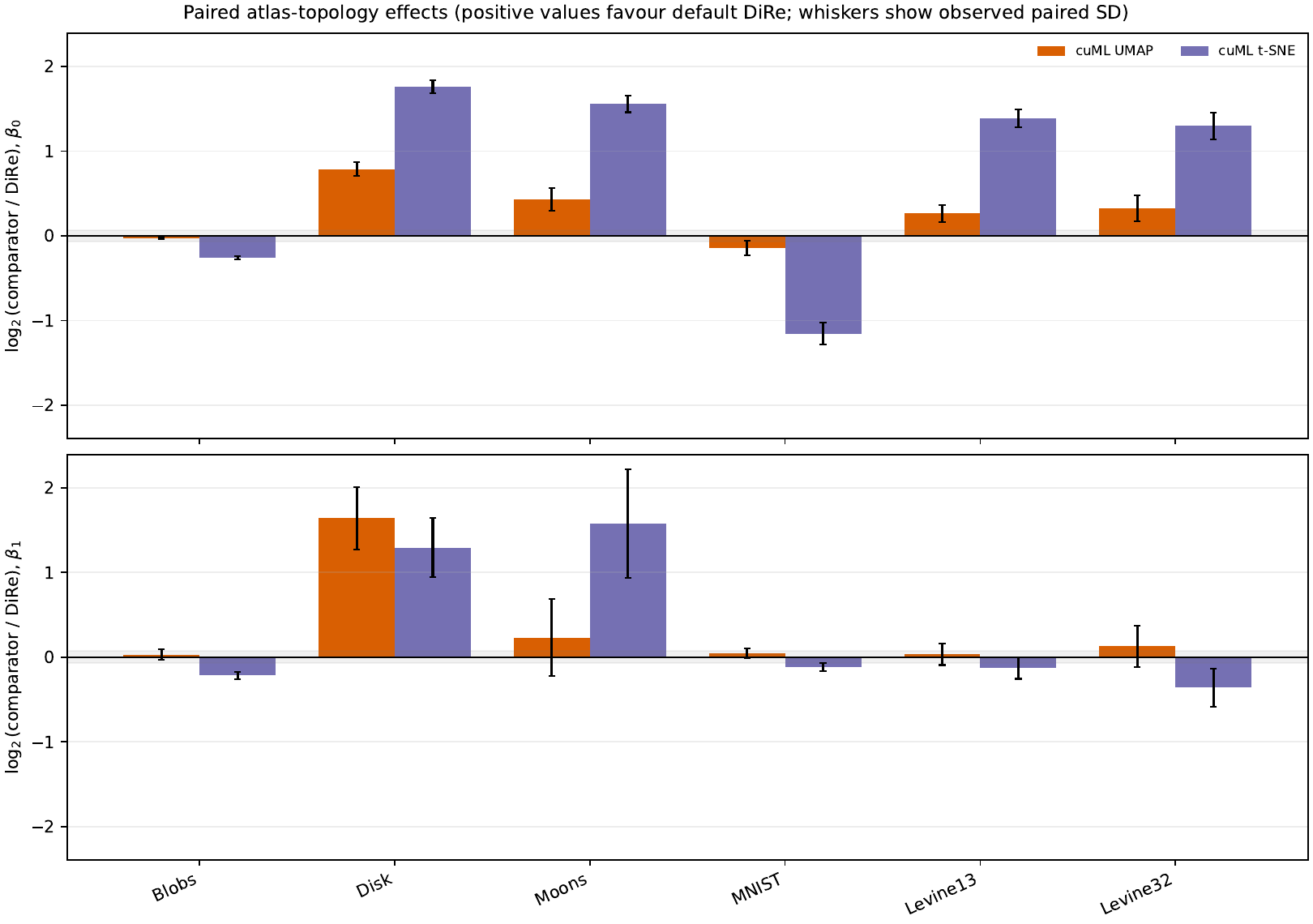}
  \caption{Paired small--suite atlas effects against default DiRe. Bars show
  the mean direction--adjusted
  $\log_2(\text{comparator discrepancy}/\text{DiRe discrepancy})$; positive
  values favour default DiRe, and whiskers show the observed sample standard
  deviation of the paired log ratios. The grey interval marks discrepancies
  within five percent. Exact raw means, minima, paired gaps, and their
  95\% descriptive mean intervals are retained in the generated CSV files.}
  \label{fig:small-atlas-topology-effects}
\end{figure}

Figures~\ref{fig:archived-quality-summary} and
\ref{fig:archived-runtime-summary}, Table~\ref{tab:small-atlas-topology}, and
Fig.~\ref{fig:small-atlas-topology-effects} replace the previous sequence of
per--metric bar charts. They show a tradeoff, not a universal winner. DiRe has
lower mean atlas discrepancy than cuML UMAP in
\SmallAtlasDireLowerVsUmapCount{} of
\SmallAtlasUmapComparisonCount{} dataset--homology comparisons. In
\SmallAtlasDireFivePctLowerVsUmapCount{} comparisons the relative advantage
exceeds the predeclared five--percent practical--difference band, and
\SmallAtlasDireIntervalLowerVsUmapCount{} paired descriptive 95\% intervals
exclude zero in DiRe's direction. Against cuML tSNE, the
\SmallAtlasTsneComparisonCount{} comparisons split evenly:
\SmallAtlasDireLowerVsTsneCount{} lower means for DiRe and
\SmallAtlasTsneLowerVsDireCount{} for tSNE. DiRe is lower in both homology
dimensions on Disk and Half--moons and in $H_0$ on both Levine datasets;
tSNE is lower in both dimensions on Blobs and MNIST and in $H_1$ on both
Levine datasets. This is a reproducible method-- and
homology--dependent distinction, not empirical redundancy.

The local metrics provide a separate tradeoff. DiRe has the highest
neighborhood overlap on Disk and is second on Half--moons, while other methods
lead that metric on the remaining datasets. In the fresh same--H100
steady--fit means, default DiRe is fastest among the three default GPU
nonlinear methods on Half--moons and Levine32 and intermediate on Blobs,
Disk, MNIST, and Levine13; it is not the slowest on any of the six datasets.
The statement that DiRe is simply ``the slowest of the three'' is therefore
not a valid summary of the reported measurements. The machine--readable cold
and steady values and regenerated detailed panels remain in the
reproducibility package.

\subsection{Million--point two--domain profile}

The large--data profile adds two public inputs from different application
domains: the released 10x embryonic mouse--brain analysis with all
1,306,127 cells, and 723,457 arXiv document embeddings. Every newly executed
reducer receives the same prepared floating--point rows in the same order. The
released Cell Ranger tSNE coordinates are retained as a separately labelled
external 10x reference, not as a same--input or runtime baseline. Dataset
revisions, download hashes, transformations, labels, method parameters,
package versions, hardware identity, failures, and fixed diagnostic row-index
subsets are recorded in the result bundle.

\begin{table}[tbp]
  \centering
  \small
  \resizebox{\textwidth}{!}{%
    \begin{tabular}{@{}lrrlll@{}}
\toprule
Dataset & $n$ & $D$ & Common input & Reference labels & License \\
\midrule
10x embryonic mouse brain & 1,306,127 & 20 & official Cell Ranger PCA scores & official k-means-20 and graph clusters & CC BY 4.0 \\
arXiv BGE-small corpus & 723,457 & 384 & mean-pooled BGE-small embeddings & primary arXiv category & CC BY 4.0 / metadata CC0 \\
\bottomrule
\end{tabular}
}
  \caption{Public large--data inputs. Here $N$ is the full number of
  observations and $D$ is the common input dimension supplied to every
  reducer.}
  \label{tab:revision3-datasets}
\end{table}

The timing result is dataset dependent but directly contradicts the claim that
DiRe is not practically viable at realistic scale. On 10x,
production--policy DiRe took \TenxDireAutoSteadySeconds{} seconds steady,
compared with \TenxCumlUmapSteadySeconds{} seconds for cuML UMAP and
\TenxCumlTsneSteadySeconds{} seconds for cuML tSNE. On arXiv the corresponding
times were \ArxivDireAutoSteadySeconds{},
\ArxivCumlUmapSteadySeconds{}, and
\ArxivCumlTsneSteadySeconds{} seconds. These are implementation measurements
on the same H100, not asymptotic equivalence claims. Production--policy DiRe
is the fastest of the three nonlinear GPU methods at both full--data
endpoints.

The quality results do not support universal dominance. On 10x, tSNE and the
released Cell Ranger layout lead most primary homology discrepancies, while
UMAP leads the labelled context and centroid--adjacency diagnostics.
Production--policy DiRe is therefore reported as fast on this dataset, not as
the topology leader. On arXiv, PCA has the lowest mean $H_0$ Betti--DTW
discrepancy. Among the primary production comparison of DiRe, cuML UMAP, and
cuML tSNE, production--policy DiRe has the lowest mean Betti--DTW discrepancy:
relative to UMAP and tSNE its mean discrepancy is lower by
\ArxivDireBetaZeroDtwReductionVsUmapPct\% and
\ArxivDireBetaZeroDtwReductionVsTsnePct\% for $H_0$, and by
\ArxivDireBetaOneDtwReductionVsUmapPct\% and
\ArxivDireBetaOneDtwReductionVsTsnePct\% for $H_1$. Table
\ref{tab:revision3-arxiv-topology-effects} reports the paired gaps and observed
standard deviations. The mean gaps are
\ArxivUmapMinusDireBetaZeroDtwGapInObservedSd{} and
\ArxivUmapMinusDireBetaOneDtwGapInObservedSd{} times their observed paired
standard deviations against UMAP, and
\ArxivTsneMinusDireBetaZeroDtwGapInObservedSd{} and
\ArxivTsneMinusDireBetaOneDtwGapInObservedSd{} times against tSNE, for $H_0$
and $H_1$, respectively. These ratios describe effect magnitude relative to
the \TopologySubsetCount{} observed paired values; they are not standard errors, $p$--values,
or claims of inferential significance. Table
\ref{tab:revision3-arxiv-topology-effects} lists every underlying
subset--specific gap rather than only its sign. In the nested subset-size
audit, the $H_1$ comparison with UMAP
reverses at the smallest nested subset $m=1000$; it holds at $m=2000$ and
$m=4000$. Exact diameter--normalised bottleneck values are nearly saturated
for arXiv and are described as non--discriminating rather than recruited as
support for the Betti--DTW result. The Q99 scale sensitivity also qualifies
the conclusion. For $H_0$, its paired UMAP-minus-DiRe gap is only
\ArxivUmapMinusDireBetaZeroQNinetyNineDtwGap{}$\pm$
\ArxivUmapMinusDireBetaZeroQNinetyNineDtwGapSd{}, while the
tSNE gap is
\ArxivTsneMinusDireBetaZeroQNinetyNineDtwGap{}$\pm$
\ArxivTsneMinusDireBetaZeroQNinetyNineDtwGapSd{}. Thus the arXiv
$H_0$ advantage is specific to the retained unit--diameter Betti--DTW
protocol, not normalization--universal. For Q99 $H_1$, the mean UMAP and tSNE
gaps are:
\ArxivUmapMinusDireBetaOneQNinetyNineDtwGap{}$\pm$
\ArxivUmapMinusDireBetaOneQNinetyNineDtwGapSd{} and
\ArxivTsneMinusDireBetaOneQNinetyNineDtwGap{}$\pm$
\ArxivTsneMinusDireBetaOneQNinetyNineDtwGapSd{}, respectively.

As an implementation--independent integrity check, the retained subsets,
diagrams, and scores were then recomputed in a CPU--only container with no GPU
exposed. This audit recomputed \TopologyAuditComparisonCount{} scalar topology
comparisons across the primary subsets, nested subset sizes, and layout seeds; all
\TopologyAuditComparisonCount{} matched the bundled values, with maximum
absolute difference \TopologyAuditMaximumAbsoluteDelta{}. The audit environment,
script hash, and machine--readable comparison records accompany the
reproduction suite.

The arXiv speed discrepancy is explained primarily by graph construction. In
the fresh paired profile, automatic IVF--PQ took
\ArxivDireAutoProfileSteadySeconds{} seconds steady, forced IVF--Flat took
\ArxivDireIvfFlatProfileSteadySeconds{} seconds, and the ratio was
\ArxivDireAutoSpeedupVsIvfFlat{}--fold. The automatic graph stage alone took
\ArxivDireAutoKnnGraphSeconds{} seconds; the recorded
\texttt{init="pca"} pathway (cuML TruncatedSVD for this 384--feature input)
and layout took \ArxivDireAutoInitializationSeconds{} and
\ArxivDireAutoLayoutSeconds{} seconds, respectively, with
\ArxivDireAutoChunkedFallbackCalls{} chunked force fallbacks. Mean fixed-query
neighbor--set overlap was \ArxivDireBackendGraphOverlap{}. The full
local/global/context/topology gate is therefore reported alongside this
speed--approximation tradeoff; the timing is not presented as evidence of
quality equivalence.

\begin{figure}[tbp]
  \centering
  \includegraphics[width=\textwidth]
    {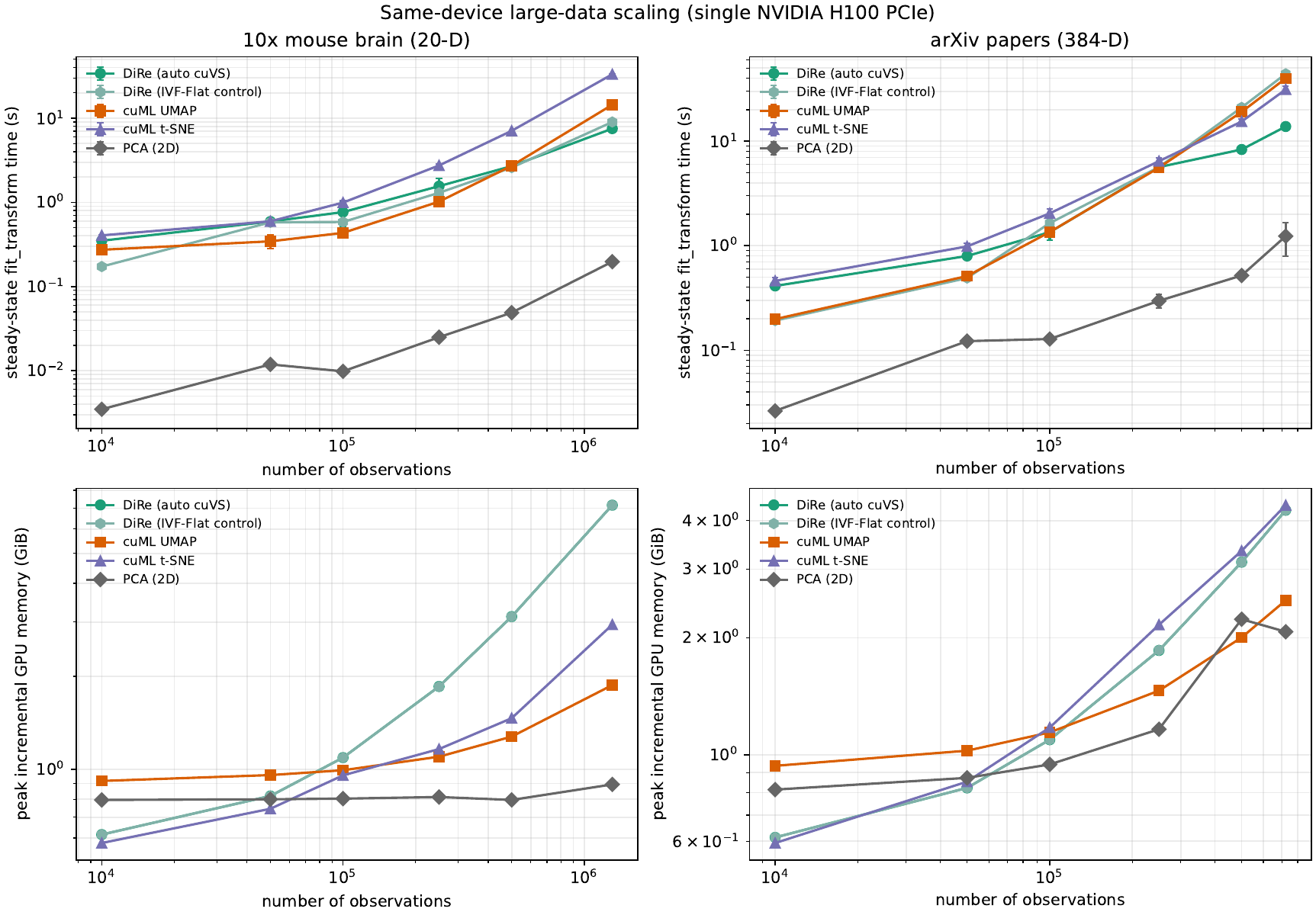}
  \caption{Same--H100 scaling through the complete 1,306,127--cell and
  723,457--document inputs. Error bars are steady--fit sample standard
  deviations; cold starts and the raw values remain in the generated table.
  CPU references are not mixed into this panel.}
  \label{fig:revision3-large-scaling}
\end{figure}

\begin{figure}[tbp]
  \centering
  \includegraphics[width=\textwidth]
    {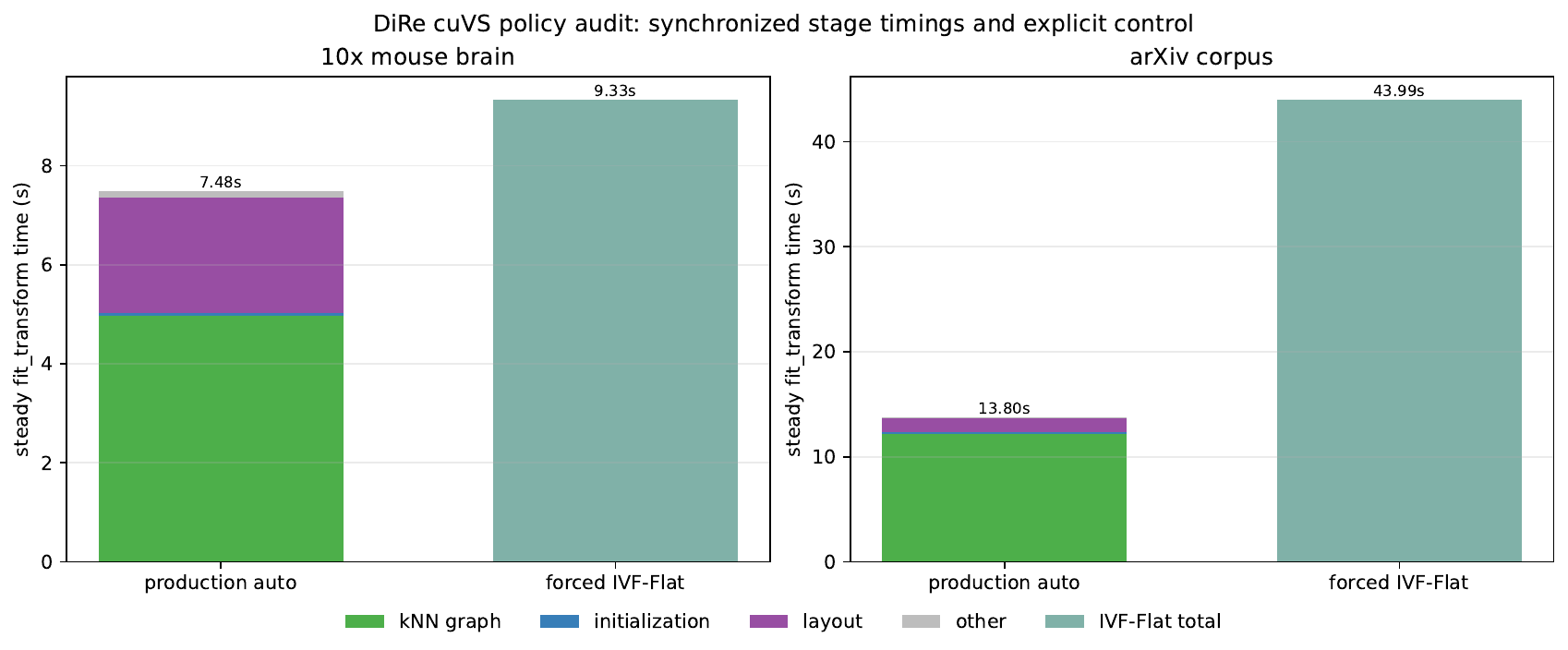}
  \caption{Paired cuVS backend--policy audit. Production
  \texttt{auto} and fresh forced IVF--Flat runs use identical inputs,
  parameters, seeds, repeat count, and H100. The forced IVF--Flat control
  remains a separately named configuration rather than being overwritten.
  Runtime is interpreted together with fixed--query graph overlap and the full
  quality comparison.}
  \label{fig:revision3-backend-policy}
\end{figure}

\begin{figure}[tbp]
  \centering
  \includegraphics[width=\textwidth]
    {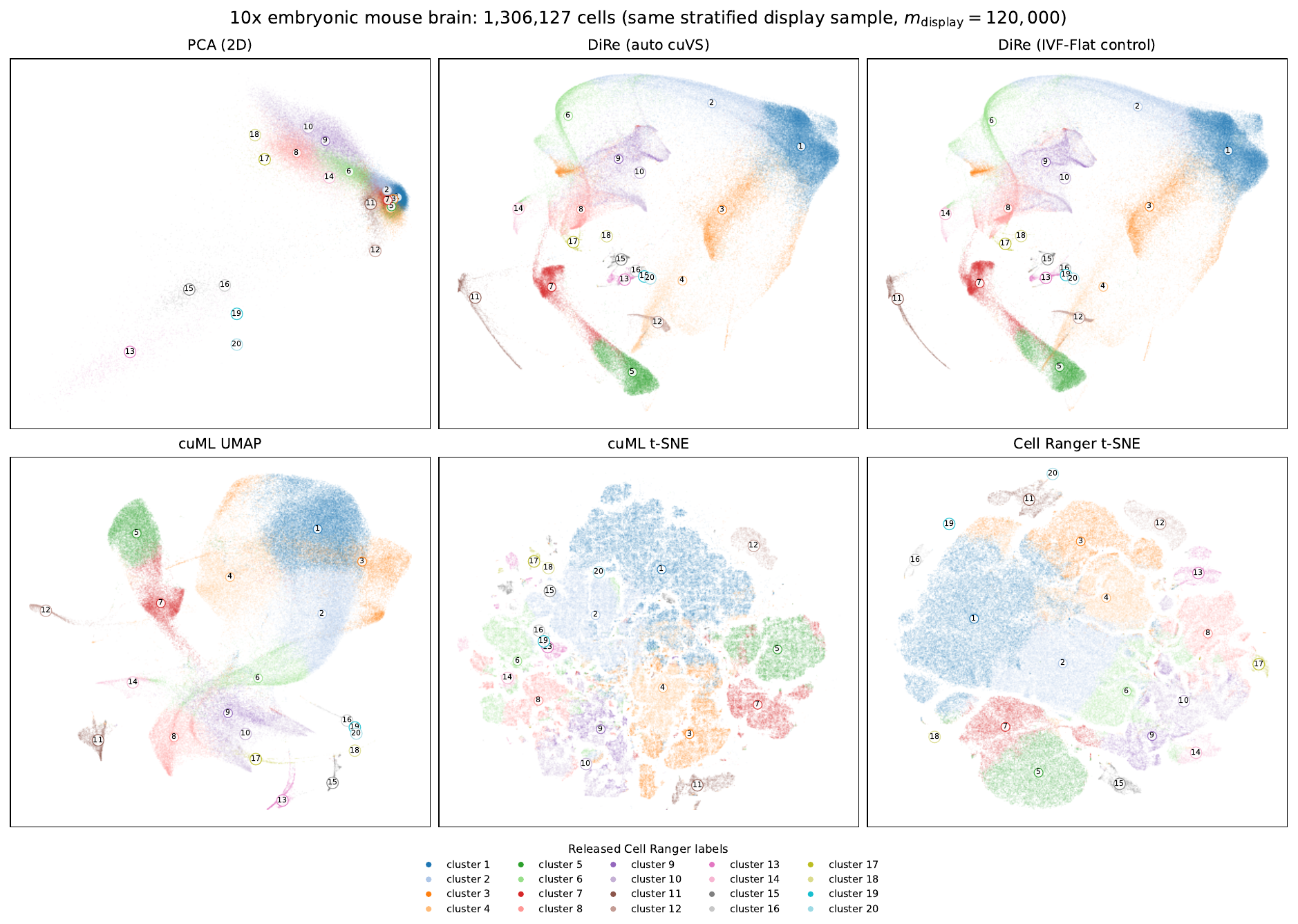}
  \caption{10x layouts on the same stratified display sample. Display sampling
  affects only rasterization; every newly run reducer is fitted to all
  $N=1,306,127$ cells, the external Cell Ranger reference supplies coordinates
  for all cells, and every quantitative metric uses its separately recorded
  protocol.}
  \label{fig:revision3-tenx-layouts}
\end{figure}

The 10x labels and marker genes are taken from the released Cell Ranger
analysis rather than inferred by visually naming clusters after reduction.
For each of the 20 official clusters, we compare its three nearest cluster
centroids in the common 20--dimensional input with its three nearest centroids
in each layout. Table~\ref{tab:revision3-tenx-adjacency} reports how many of
those 60 directed relationships are preserved. The corresponding pair-level
CSV attaches the source cluster's released positive markers to every
preserved, lost, and introduced relationship. This makes visual interpretation
auditable while avoiding the circular claim that stronger apparent separation
must be biologically more faithful. The complete released marker list is
given in Table~\ref{tab:revision3-tenx-markers}.

Several preserved input relationships are directly interpretable from those
released markers. Production--policy DiRe retains both directions between
clusters 9 and 10, whose strongest markers include the cell--cycle genes
\textit{Rrm2}, \textit{Hmgb2}, \textit{Ube2c}, \textit{Cenpf}, and
\textit{Ccnb1}; it retains 20$\rightarrow$19, whose source and target both
include \textit{S100a8}; 13$\rightarrow$15, linking the
\textit{Cldn5}/\textit{Flt1} and \textit{Rgs5}/\textit{Ndufa4l2} marker
profiles; and 19$\rightarrow$16, linking two marker profiles with myeloid
genes. These are examples of input relationships retained in the benchmark
layout configured with $d=2$, not discoveries made from the picture. The aggregate
result is also important: production--policy DiRe preserves
\TenxDireAutoAdjacencyPreserved{} of 60 directed input relationships and
introduces \TenxDireAutoAdjacencyIntroduced{}, whereas UMAP preserves
\TenxCumlUmapAdjacencyPreserved{} and introduces
\TenxCumlUmapAdjacencyIntroduced{}. We therefore do not use the examples to claim biological or
metric superiority. An independent recomputation from every bundled full
layout reproduces all stored centroid adjacencies exactly. The primary
equal--cluster recall gives each released cluster the same weight; the
separately reported source--cell--weighted recalls are
\TenxDireAutoAdjacencyCellWeighted{} for DiRe and
\TenxCumlUmapAdjacencyCellWeighted{} for UMAP. Excluding the three source
clusters with fewer than 1,000 cells does not explain the difference:
equal--cluster recall becomes \TenxDireAutoAdjacencyFilteredEqual{} and
\TenxCumlUmapAdjacencyFilteredEqual{}, respectively. The comparison is also
$k$--dependent: for $k=1,2,3$, respectively, DiRe's equal--cluster recalls are
\TenxDireAutoAdjacencyKOne{}, \TenxDireAutoAdjacencyKTwo{}, and
\TenxDireAutoAdjacencyKThree{}, while UMAP's are
\TenxCumlUmapAdjacencyKOne{}, \TenxCumlUmapAdjacencyKTwo{}, and
\TenxCumlUmapAdjacencyKThree{}. These sensitivity records accompany the
pair--level CSV and are not used to replace the predeclared three--centroid
result. The one--factor DiRe controls also show that the global arrangement is
initialization-- and configuration--dependent rather than a corrupted metric:
spectral--initialised DiRe preserves
\TenxDireSpectralAdjacencyPreserved{} relationships with source--cell--weighted
recall \TenxDireSpectralAdjacencyCellWeighted{}. UMAP remains the leader in the
predeclared equal--cluster comparison, and the controls are reported as
sensitivity results rather than substituted for production--policy DiRe.

\begin{table}[tbp]
  \centering
  \resizebox{\textwidth}{!}{%
    \begin{tabular}{@{}lrrrrr@{}}
\toprule
Method & Input directed edges & Preserved & Introduced & Equal-cluster recall & Cell-weighted recall \\
\midrule
PCA (2D) & 60 & 26 & 34 & 0.433 & 0.401 \\
DiRe (auto cuVS) & 60 & 18 & 42 & 0.300 & 0.328 \\
DiRe (IVF-Flat control) & 60 & 17 & 43 & 0.283 & 0.315 \\
DiRe (spectral init) & 60 & 26 & 34 & 0.433 & 0.688 \\
cuML UMAP & 60 & 29 & 31 & 0.483 & 0.719 \\
cuML t-SNE & 60 & 21 & 39 & 0.350 & 0.346 \\
Cell Ranger t-SNE & 60 & 21 & 39 & 0.350 & 0.409 \\
\bottomrule
\end{tabular}
}
  \caption{Preservation of directed three-nearest-centroid relationships
  between the common 20--dimensional 10x input and each benchmark layout
  configured with $d=2$. An introduced edge displaces an input edge; it is not
  interpreted as a newly discovered biological relation. Equal--cluster recall
  weights each of the 20 released clusters equally; source--cell--weighted
  recall weights each source cluster by its released cell count.}
  \label{tab:revision3-tenx-adjacency}
\end{table}

\begin{figure}[tbp]
  \centering
  \includegraphics[width=\textwidth]
    {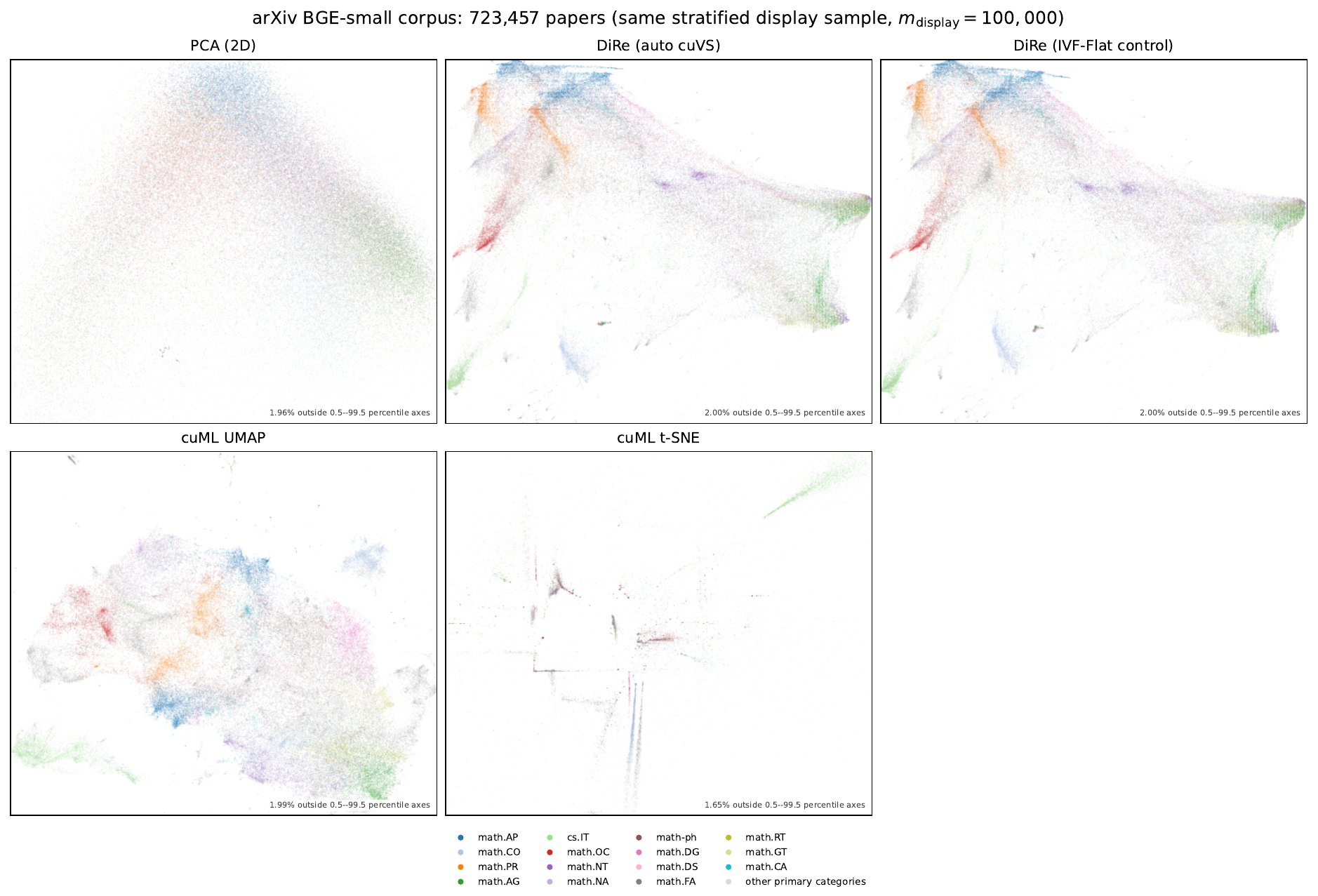}
  \caption{arXiv layouts on the same stratified display sample. All
  $N=723,457$ document embeddings are fitted; the slight per--panel display
  clipping and clipped fraction are stated in the figure and are not applied
  to any metric.}
  \label{fig:revision3-arxiv-layouts}
\end{figure}

\begin{figure}[tbp]
  \centering
  \includegraphics[width=\textwidth]
    {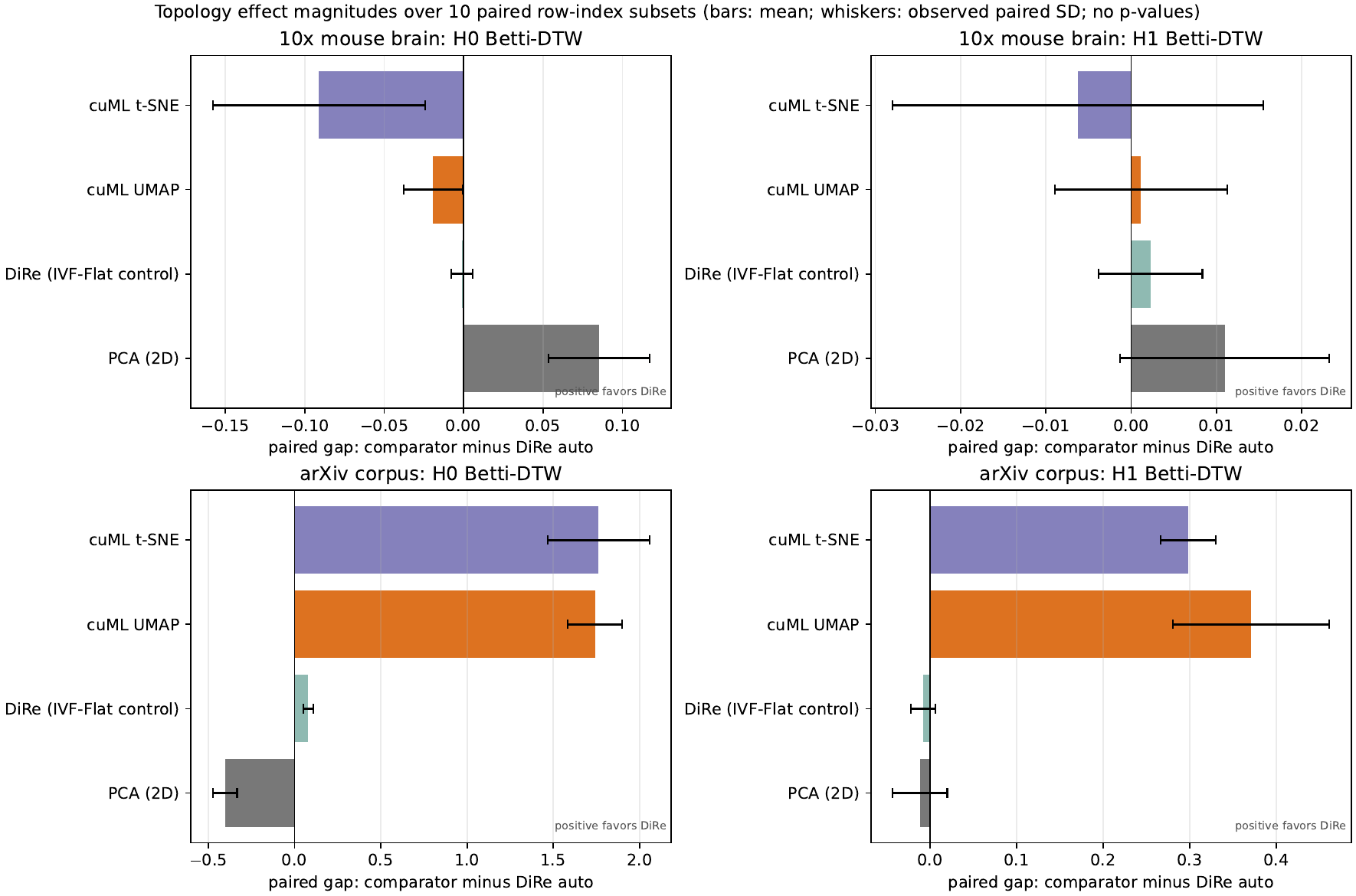}
  \caption{Paired topology effect magnitudes. Each bar is the mean of
  comparator discrepancy minus production--policy DiRe discrepancy over
  \TopologySubsetCount{} independently seeded, shared 4,000--row topology
  subsets; positive
  values favour DiRe, and whiskers are the observed standard deviation of the
  paired subset--specific gaps. Each subset uses identical row indices
  for the high--dimensional reference and every embedding. These are descriptive
  robustness records, not $p$--values or claims of conventional statistical
  significance.}
  \label{fig:revision3-topology-effects}
\end{figure}

\begin{table}[tbp]
  \centering
  \resizebox{\textwidth}{!}{%
    \begin{tabular}{@{}llrrrrrrrr@{}}
\toprule
Dataset & Effective auto index & Cold (s) & Steady (s) & Graph (s) & PCA (s) & Layout (s) & IVF-Flat (s) & Speed-up & Graph overlap \\
\midrule
10x & ivf\_pq & 10.3 & 7.48 & 4.96 & 0.061 & 2.33 & 9.33 & 1.25$\times$ & 0.626 \\
arXiv & ivf\_pq & 14.6 & 13.8 & 12.2 & 0.16 & 1.30 & 44.0 & 3.19$\times$ & 0.840 \\
\bottomrule
\end{tabular}
}
  \caption{Cold and steady backend timing decomposition, fixed--query graph
  overlap, and the fresh forced--IVF--Flat/production--auto speed ratio.}
  \label{tab:revision3-backend-policy}
\end{table}

\begin{table}[tbp]
  \centering
  \resizebox{\textwidth}{!}{%
    \begin{tabular}{@{}llrrrrr@{}}
\toprule
Metric & Comparator & DiRe auto & Comparator & Paired gap $\pm$ SD & Relative gap & 10 paired gaps \\
\midrule
$\beta_0$ DTW & cuML t-SNE & 3.5741 & 5.3357 & 1.7616$\pm$0.2948 & 33.0\% & \shortstack[l]{+2.0907, +1.4510, +1.9125, +1.7613, +1.3157\\+2.1270, +1.8870, +1.8897, +1.3340, +1.8467} \\
$\beta_1$ DTW & cuML t-SNE & 0.1467 & 0.4447 & 0.2980$\pm$0.0316 & 67.0\% & \shortstack[l]{+0.2833, +0.2908, +0.2770, +0.3445, +0.2655\\+0.3047, +0.2518, +0.3097, +0.3500, +0.3030} \\
$\beta_0$ Q99--DTW & cuML t-SNE & 3.6111 & 2.5703 & -1.0407$\pm$0.9596 & -40.5\% & \shortstack[l]{-2.2800, +0.7255, -0.9315, -1.7763, -0.1195\\-1.6362, -1.6250, -1.7570, -0.0317, -0.9755} \\
$\beta_1$ Q99--DTW & cuML t-SNE & 0.1461 & 0.2385 & 0.0924$\pm$0.1154 & 38.7\% & \shortstack[l]{-0.0350, +0.2652, +0.0302, +0.0340, +0.2023\\+0.0255, +0.0155, +0.0025, +0.2815, +0.1020} \\
$\beta_0$ DTW & cuML UMAP & 3.5741 & 5.3161 & 1.7421$\pm$0.1581 & 32.8\% & \shortstack[l]{+1.8002, +1.7030, +1.8518, +1.8477, +1.6282\\+1.9605, +1.7803, +1.7347, +1.3753, +1.7388} \\
$\beta_1$ DTW & cuML UMAP & 0.1467 & 0.5173 & 0.3706$\pm$0.0904 & 71.6\% & \shortstack[l]{+0.1973, +0.3882, +0.3738, +0.4700, +0.3890\\+0.2317, +0.3568, +0.4160, +0.4668, +0.4167} \\
$\beta_0$ Q99--DTW & cuML UMAP & 3.6111 & 3.6223 & 0.0113$\pm$0.0591 & 0.3\% & \shortstack[l]{-0.0520, +0.0335, +0.0072, -0.0693, -0.0368\\+0.0605, +0.0932, +0.1002, -0.0160, -0.0080} \\
$\beta_1$ Q99--DTW & cuML UMAP & 0.1461 & 0.2360 & 0.0899$\pm$0.1174 & 38.1\% & \shortstack[l]{+0.0200, +0.0353, +0.0152, +0.1187, +0.4103\\+0.0780, +0.0487, +0.0682, +0.0830, +0.0212} \\
$\beta_0$ DTW & DiRe (IVF-Flat control) & 3.5741 & 3.6531 & 0.0791$\pm$0.0285 & 2.2\% & \shortstack[l]{+0.0192, +0.0968, +0.0725, +0.0947, +0.1035\\+0.1040, +0.0860, +0.1040, +0.0608, +0.0490} \\
$\beta_1$ DTW & DiRe (IVF-Flat control) & 0.1467 & 0.1385 & -0.0082$\pm$0.0143 & -5.9\% & \shortstack[l]{-0.0237, -0.0088, -0.0143, -0.0195, +0.0255\\-0.0195, -0.0030, -0.0030, -0.0155, -0.0005} \\
$\beta_0$ DTW & PCA (2D) & 3.5741 & 3.1713 & -0.4027$\pm$0.0704 & -12.7\% & \shortstack[l]{-0.4290, -0.3980, -0.3382, -0.3510, -0.3775\\-0.2700, -0.4397, -0.5088, -0.4660, -0.4490} \\
$\beta_1$ DTW & PCA (2D) & 0.1467 & 0.1350 & -0.0117$\pm$0.0314 & -8.6\% & \shortstack[l]{-0.0205, -0.0435, -0.0473, +0.0502, -0.0268\\-0.0003, -0.0318, +0.0310, -0.0212, -0.0065} \\
\bottomrule
\end{tabular}
}
  \caption{arXiv paired Betti--DTW effects under the retained unit--diameter
  protocol and the separately labelled Q99 scale sensitivity. The gap is
  comparator minus production--policy DiRe, so a positive gap favours DiRe.
  Each topology subset is one independently seeded set of $m=4000$ row indices
  applied identically to the high--dimensional reference and every embedding.
  The final column reports every subset--specific paired gap in fixed seed
  order rather than reducing the signs to a count.}
  \label{tab:revision3-arxiv-topology-effects}
\end{table}

\subsection{Dataset: Blobs}

\begin{figure}[h]
    \centering
    \begin{subfigure}[b]{0.45\textwidth}
        \centering
        \includegraphics[width=\textwidth]{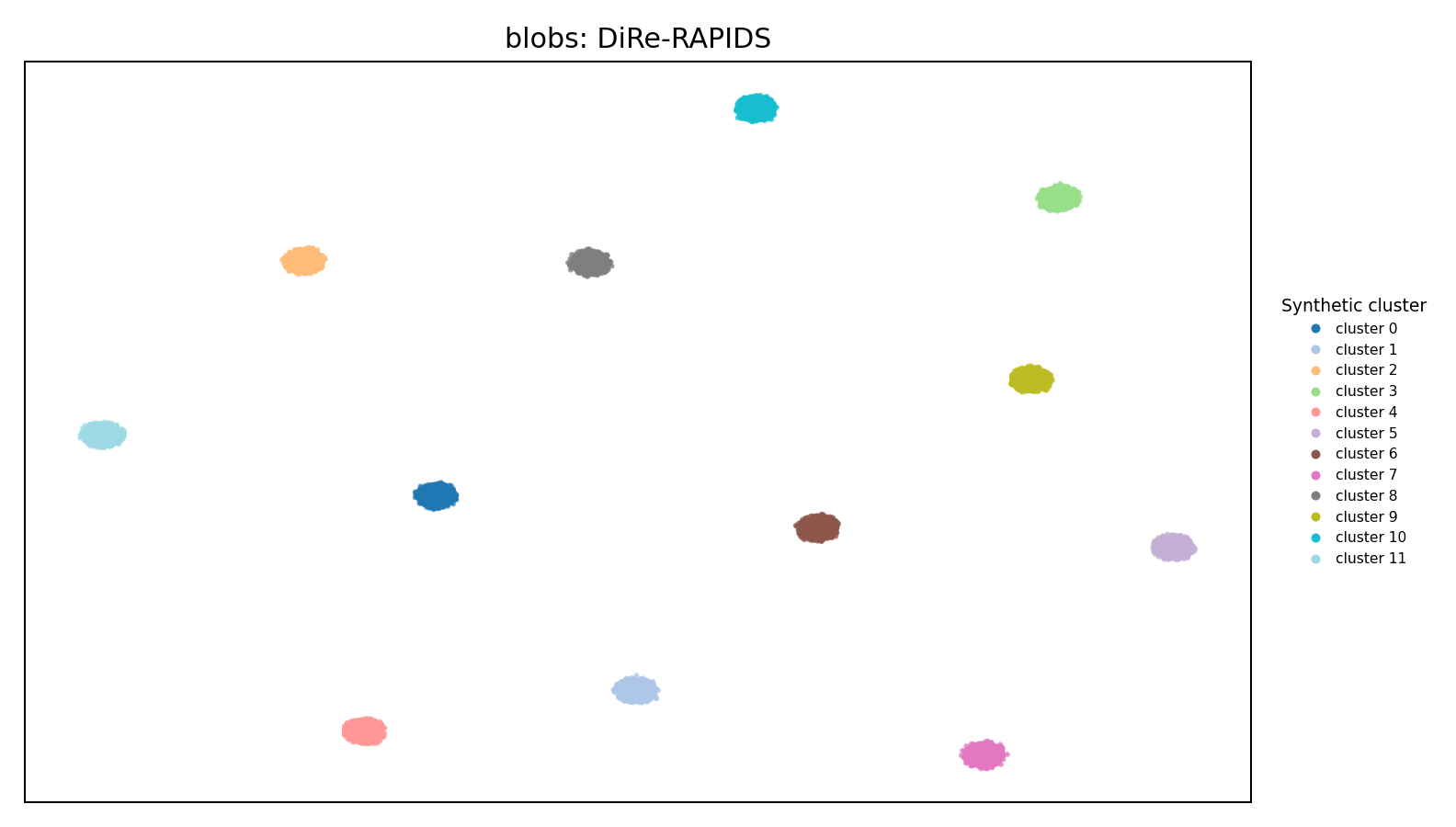}
        \caption{DiRe--RAPIDS}
        \label{fig:blobs-dire-rapids}
    \end{subfigure}
    \hfill
    \begin{subfigure}[b]{0.45\textwidth}
        \centering
        \includegraphics[width=\textwidth]{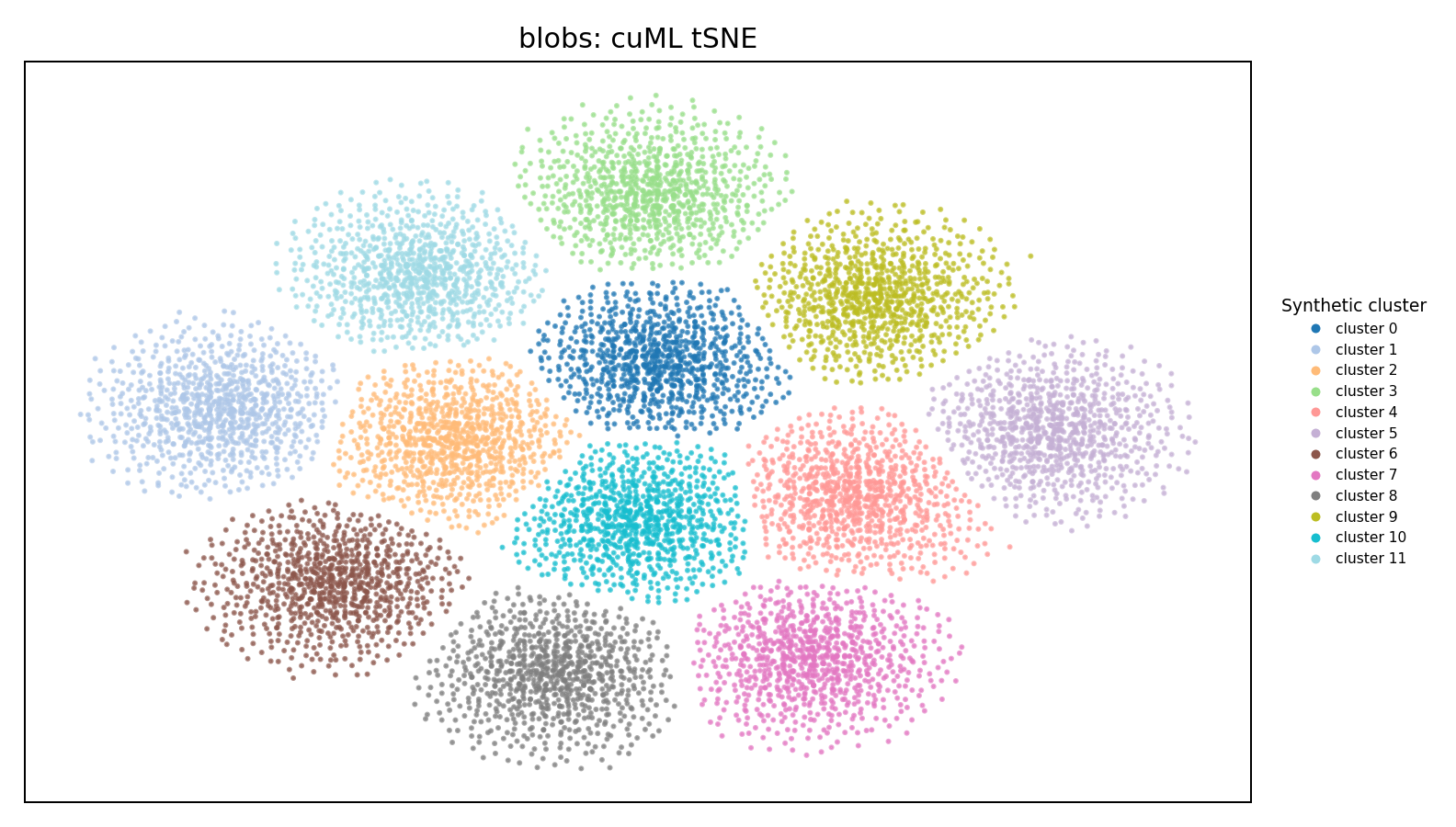}
        \caption{cuML tSNE}
        \label{fig:blobs-tsne}
    \end{subfigure}
    \vspace{1em}
    \begin{subfigure}[b]{0.45\textwidth}
        \centering
        \includegraphics[width=\textwidth]{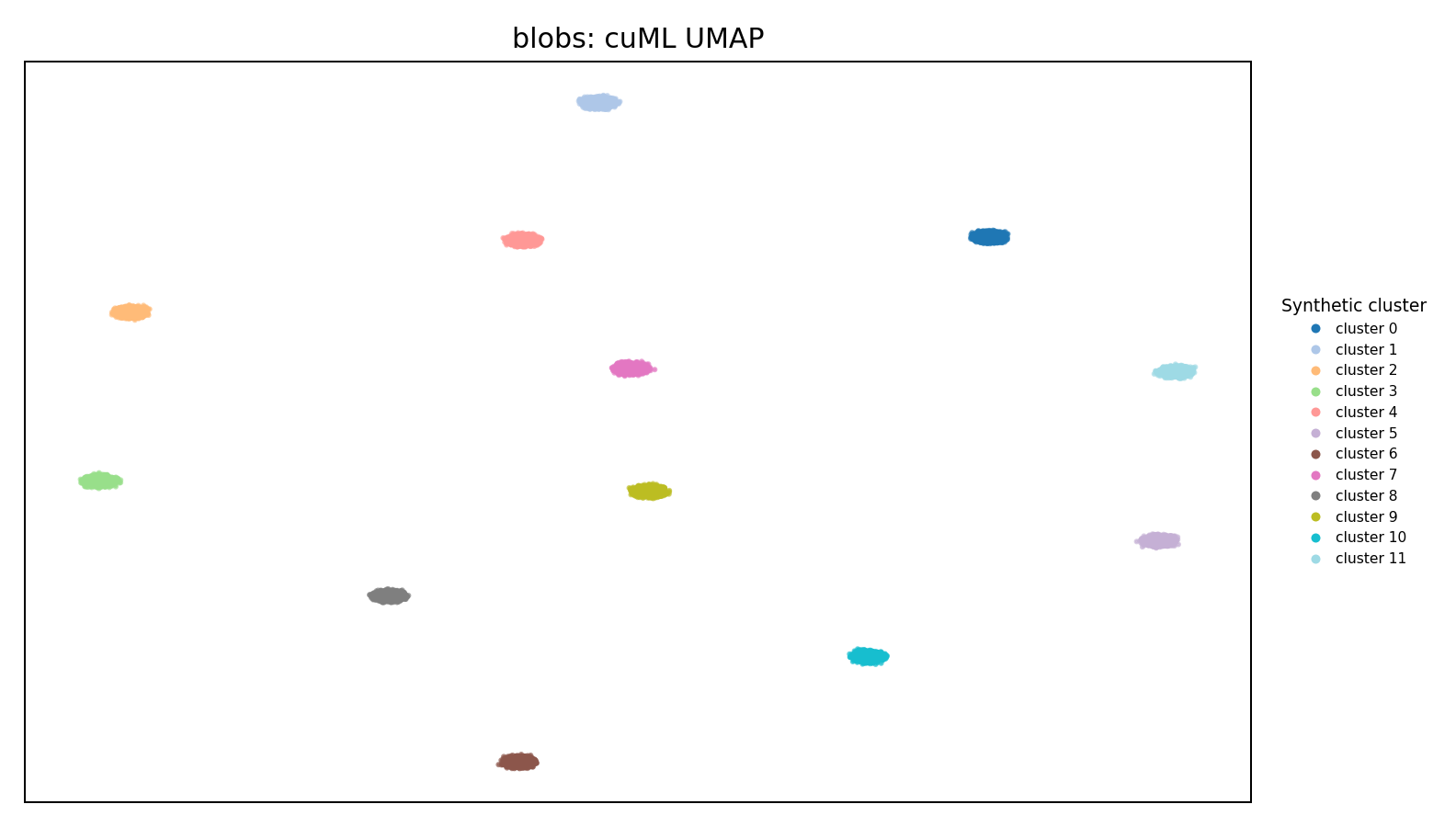}
        \caption{cuML UMAP}
        \label{fig:blobs-cuml-umap}
    \end{subfigure}
    \hfill
    \begin{subfigure}[b]{0.45\textwidth}
        \centering
        \includegraphics[width=\textwidth]{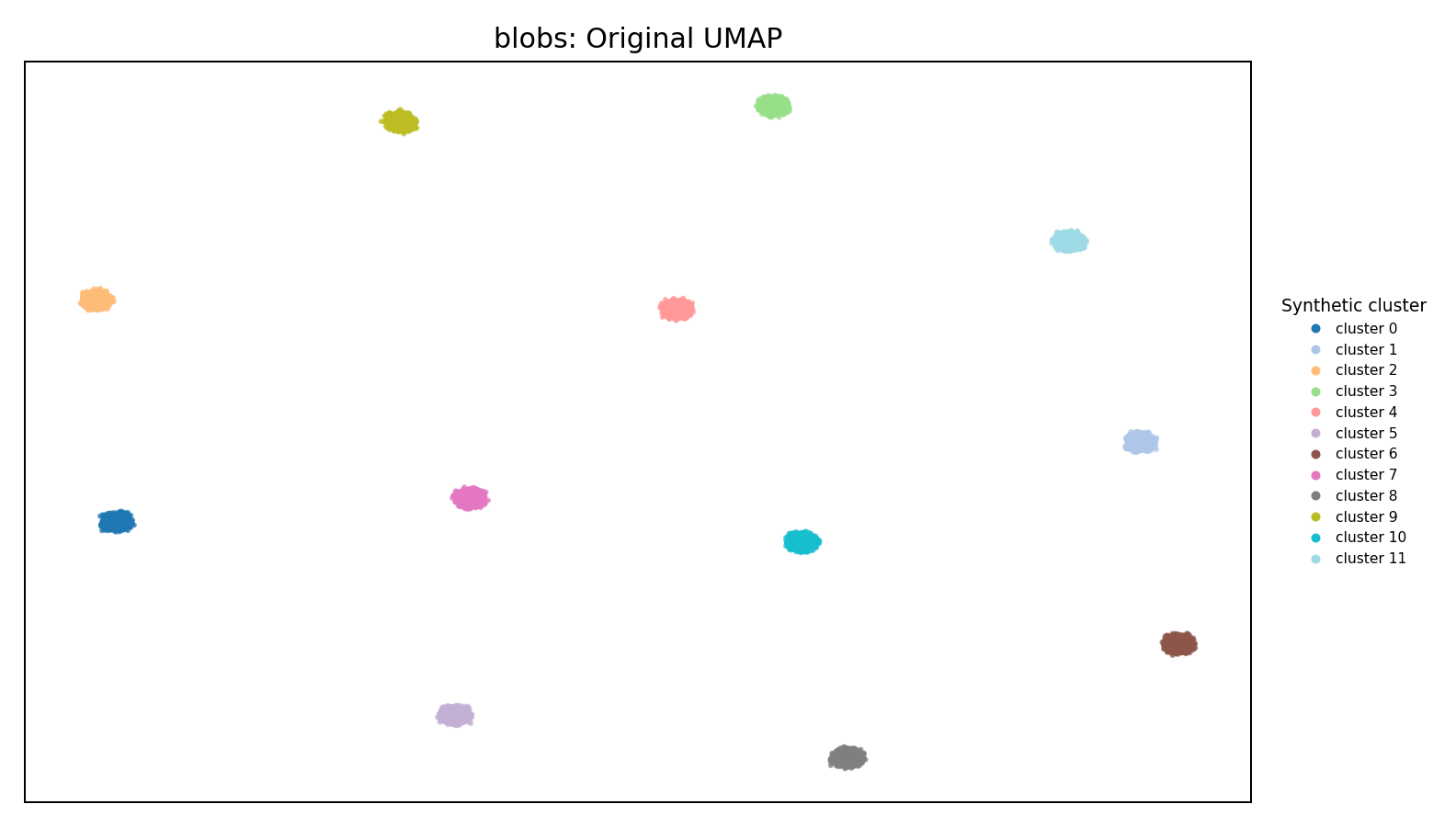}
        \caption{Original UMAP}
        \label{fig:blobs-umap}
    \end{subfigure}
    \caption{"Blobs" dataset embeddings}
    \label{fig:blobs}
\end{figure}

The benchmark embeddings configured with $d=2$ for the blobs dataset are given in Fig.~\ref{fig:blobs}. The dataset has $10$K points, 12 Gaussian clusters, and ambient dimension $1000$. It is primarily a sanity check: under appropriate parameters, all methods should recover the cluster structure.

The key metrics are summarized in
Fig.~\ref{fig:archived-quality-summary}; the corresponding raw values and
detailed generated panels are retained in the reproducibility package. We
compare the visual embedding, local metrics, and persistence metrics, since
these criteria capture different aspects of the output.

The local metrics can rank the methods differently from the global metrics, which is expected because stress and neighborhood preservation measure local metric distortion rather than cluster--level topology.

The average neighborhood preservation score should not be interpreted as a complete quality measure. Even simple dimension reductions can change nearest--neighbor identities while preserving larger clusters or connected components. For example, the planar point cloud $X_k = \{x_n = (1/n, n) : n=1,\ldots,k\}$ has $x_k$ as the point farthest from the origin, while its projection to the horizontal axis is closest to $0$.

\subsection{Dataset: MNIST Digits}

MNIST Digits provides a labeled image benchmark with well known class structure. This benchmark compares whether the methods preserve digit classes, transitions between visually similar digits, and the persistent homology summaries of the high--dimensional data.

The key metrics are included in Fig.~\ref{fig:archived-quality-summary}. In
this dataset, visual cluster separation, classifier context, and persistence
distances do not induce the same ranking, so their raw records are also
reported separately in the reproduction archive.

Even when all methods reasonably preserve labeled context, the SVM and kNN context scores can distinguish linear separability from local class consistency.

\begin{figure}[h]
    \centering
    \begin{subfigure}[b]{0.45\textwidth}
        \centering
        \includegraphics[width=\textwidth]{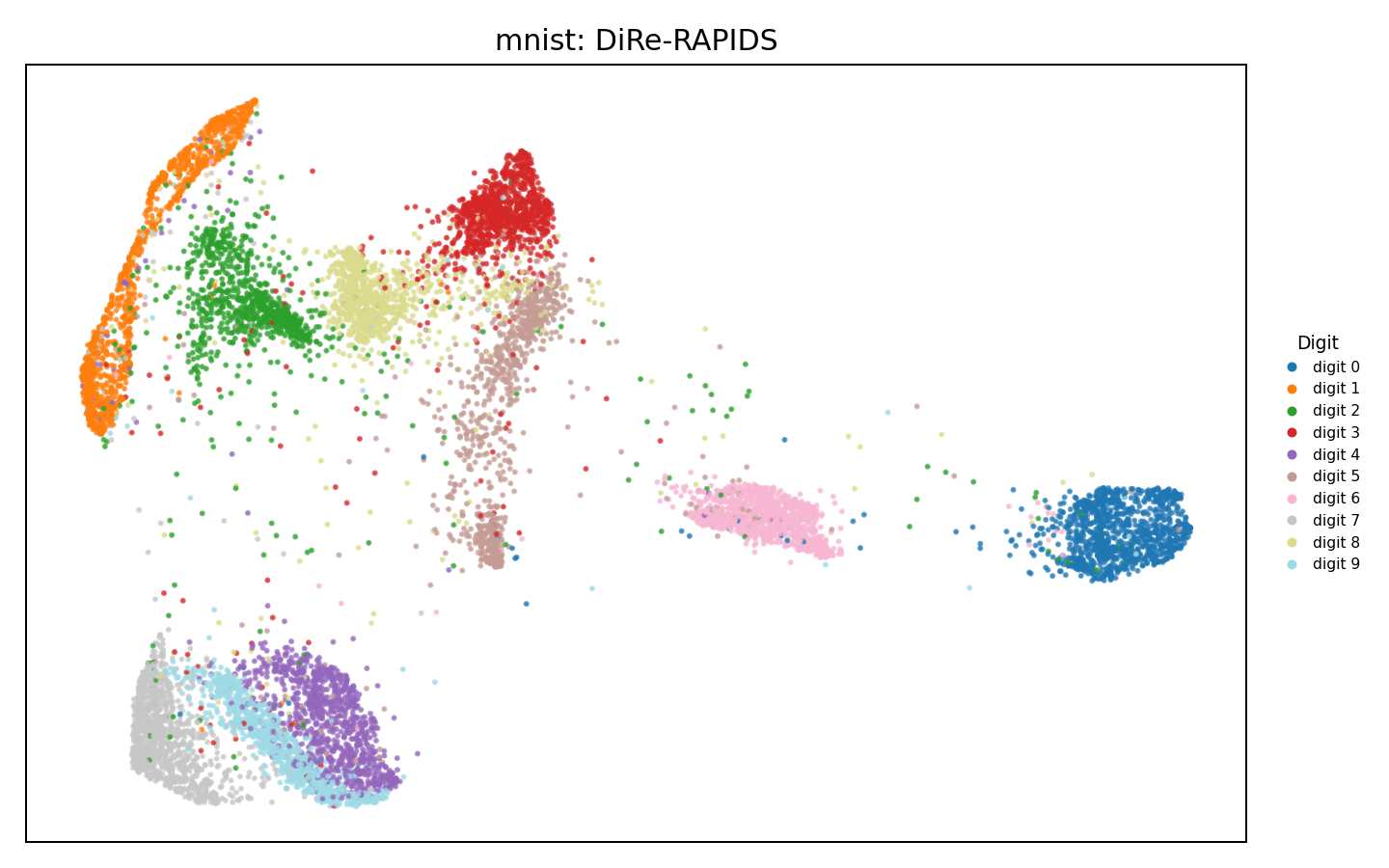}
        \caption{DiRe--RAPIDS}
        \label{fig:mnist-dire-rapids}
    \end{subfigure}
    \hfill
    \begin{subfigure}[b]{0.45\textwidth}
        \centering
        \includegraphics[width=\textwidth]{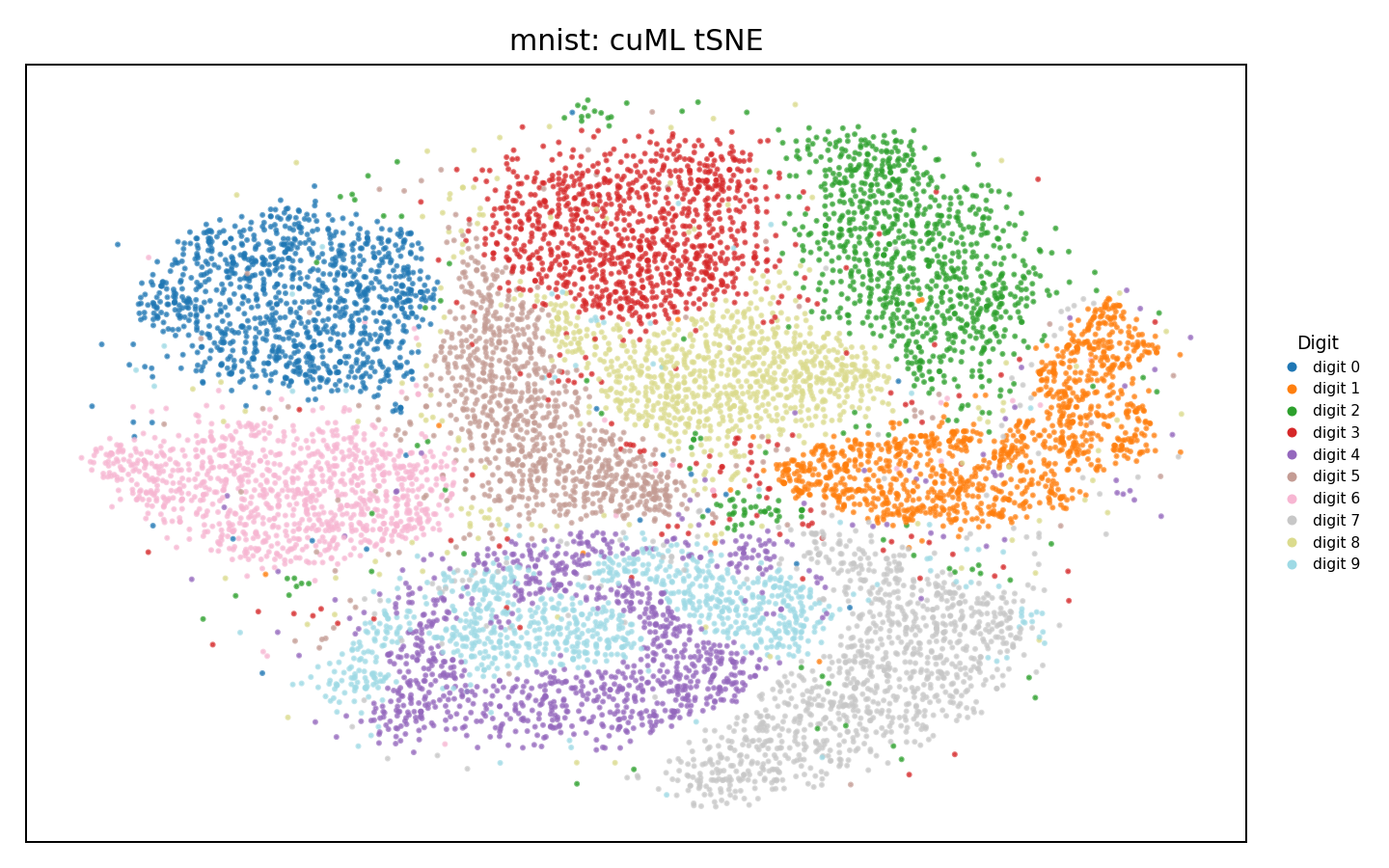}
        \caption{cuML tSNE}
        \label{fig:mnist-tsne}
    \end{subfigure}
    \vspace{1em}
    \begin{subfigure}[b]{0.45\textwidth}
        \centering
        \includegraphics[width=\textwidth]{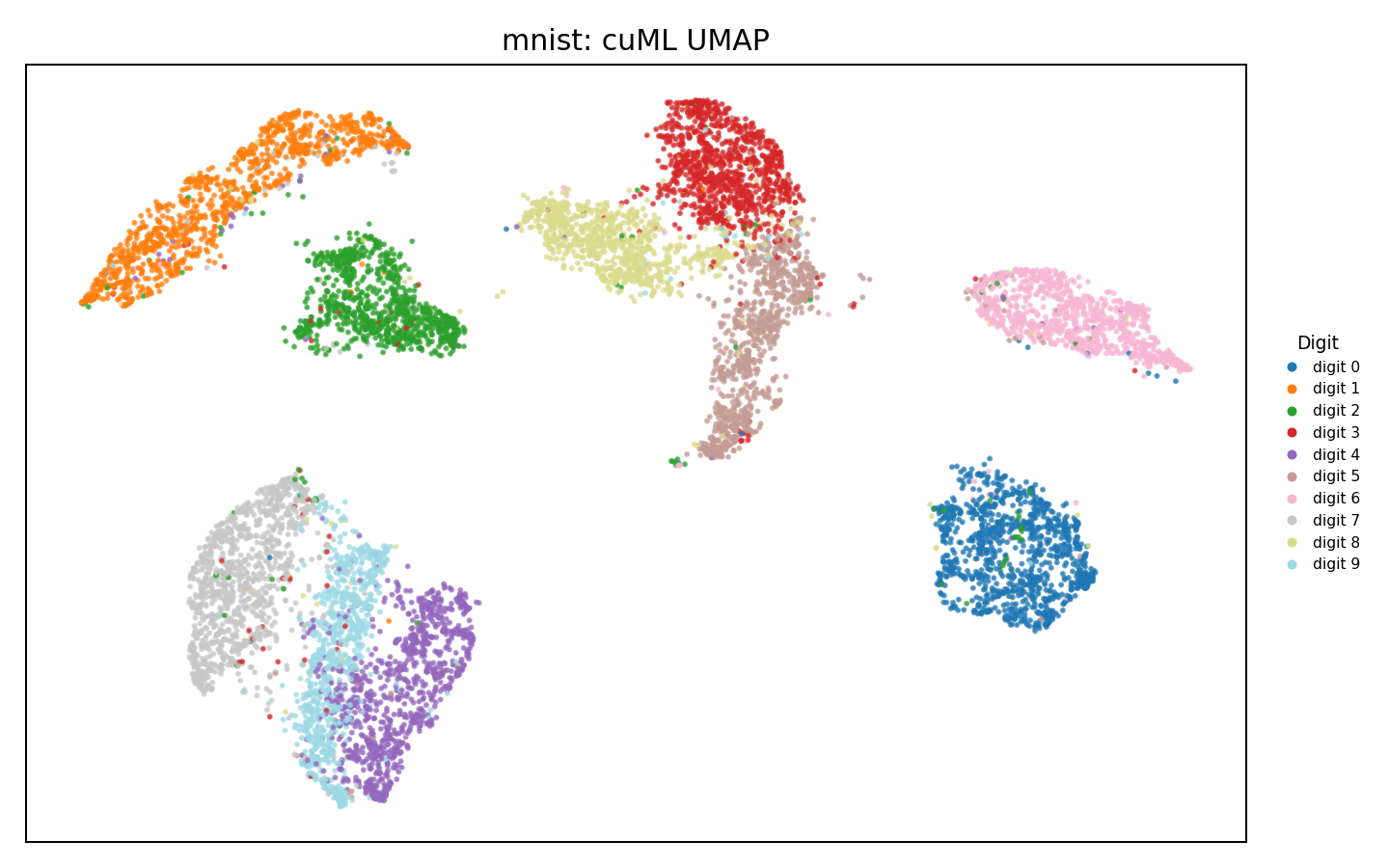}
        \caption{cuML UMAP}
        \label{fig:mnist-cuml-umap}
    \end{subfigure}
    \hfill
    \begin{subfigure}[b]{0.45\textwidth}
        \centering
        \includegraphics[width=\textwidth]{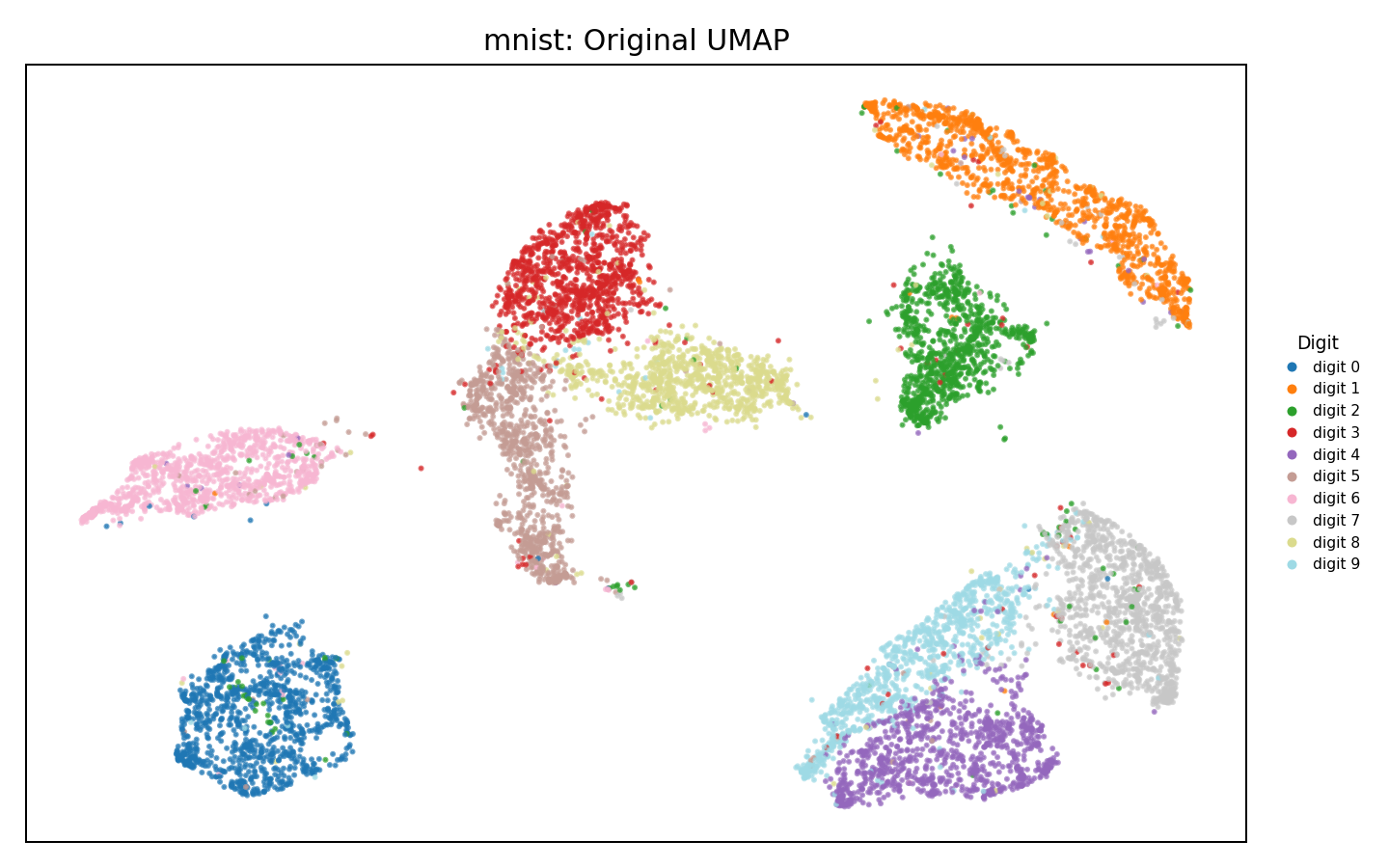}
        \caption{Original UMAP}
        \label{fig:mnist-umap}
    \end{subfigure}
    \caption{"MNIST Digits" dataset embeddings}
    \label{fig:mnist}
\end{figure} 

\subsection{Dataset: Disk Uniform}

Here we embed a 2D disk with points uniformly distributed in it. Since the input already has a simple low--dimensional geometry, this benchmark tests whether a method unnecessarily distorts connectedness, density, or boundary shape.

A similar family of tests can be run on a sphere and an ellipsoid in 3D with uniformly distributed points. See the DiRe--RAPIDS repository for additional datasets \cite{dire-rapids}.

The metrics are included in Fig.~\ref{fig:archived-quality-summary}. We treat
persistence metrics and local distortion metrics as complementary: a method
may preserve nearest neighbors while changing the global shape of this simple
input.

\begin{figure}[ht]
    \centering
    \begin{subfigure}[b]{0.45\textwidth}
        \centering
        \includegraphics[width=\textwidth]{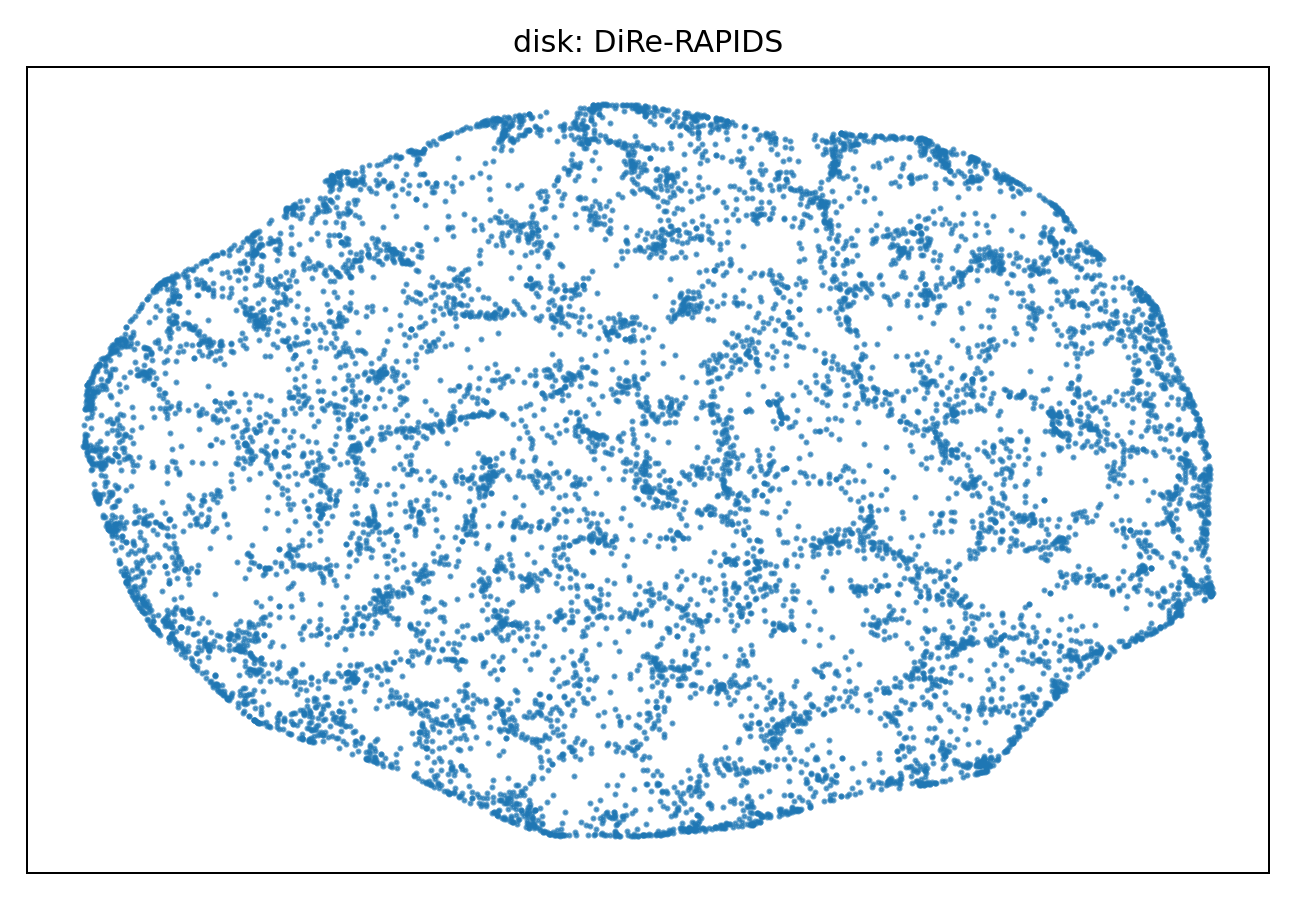}
        \caption{DiRe--RAPIDS}
        \label{fig:disk-dire-rapids}
    \end{subfigure}
    \hfill
    \begin{subfigure}[b]{0.45\textwidth}
        \centering
        \includegraphics[width=\textwidth]{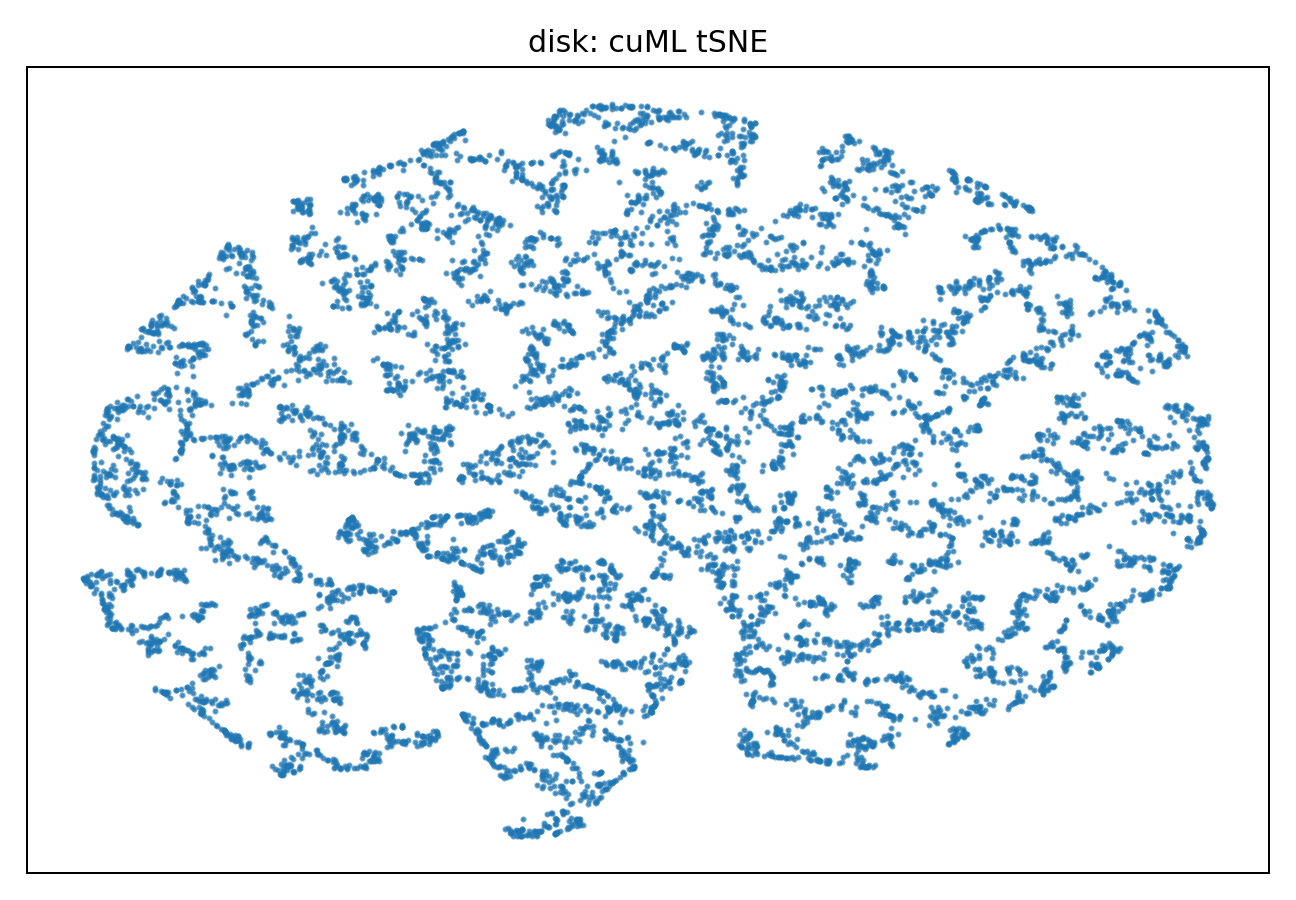}
        \caption{cuML tSNE}
        \label{fig:disk-tsne}
    \end{subfigure}
    \vspace{1em}
    \begin{subfigure}[b]{0.45\textwidth}
        \centering
        \includegraphics[width=\textwidth]{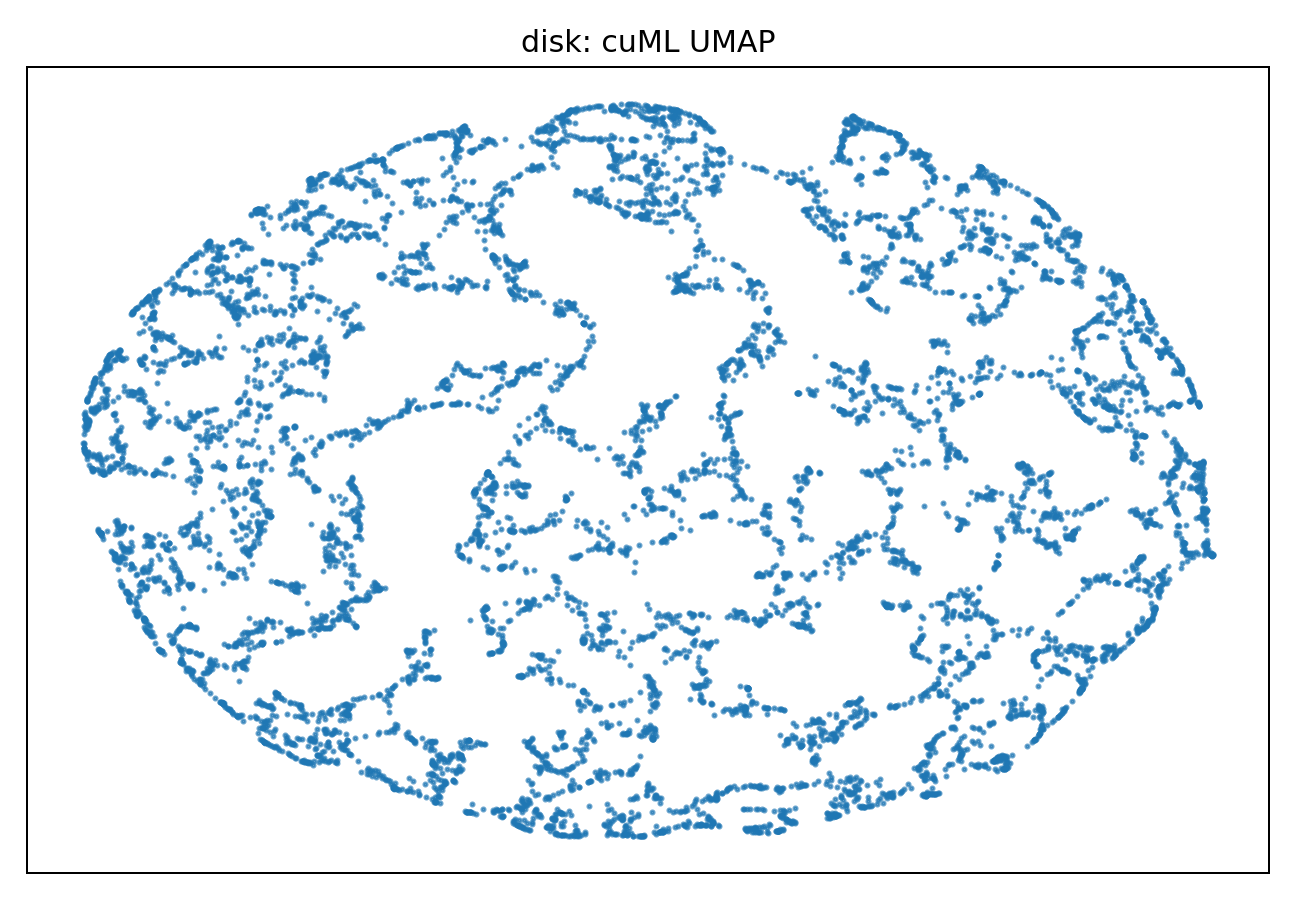}
        \caption{cuML UMAP}
        \label{fig:disk-cuml-umap}
    \end{subfigure}
    \hfill
    \begin{subfigure}[b]{0.45\textwidth}
        \centering
        \includegraphics[width=\textwidth]{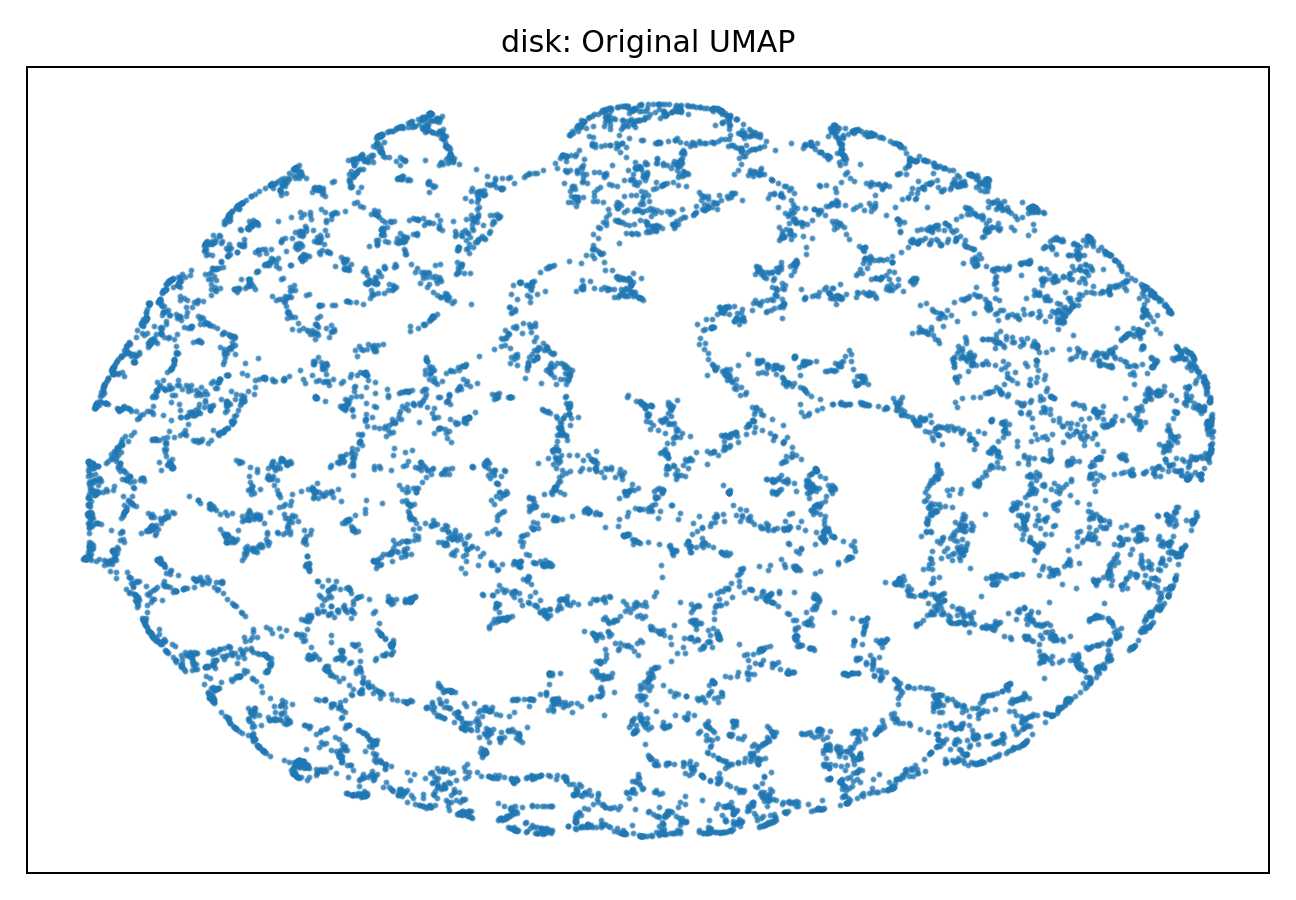}
        \caption{Original UMAP}
        \label{fig:disk-umap}
    \end{subfigure}
    \caption{"Disk Uniform" dataset embeddings}
    \label{fig:disk}
\end{figure}

\subsection{Dataset: Half-moons}

The half--moons dataset tests whether a method preserves the non--linear
relation between two labeled components. Since the input classes are not
linearly separable, a benchmark embedding configured with $d=2$ that makes them linearly
separable may improve a downstream classifier while changing the original
context. The context preservation score is included in
Fig.~\ref{fig:archived-quality-summary}.

\begin{figure}[ht]
    \centering
    \begin{subfigure}[b]{0.45\textwidth}
        \centering
        \includegraphics[width=\textwidth]{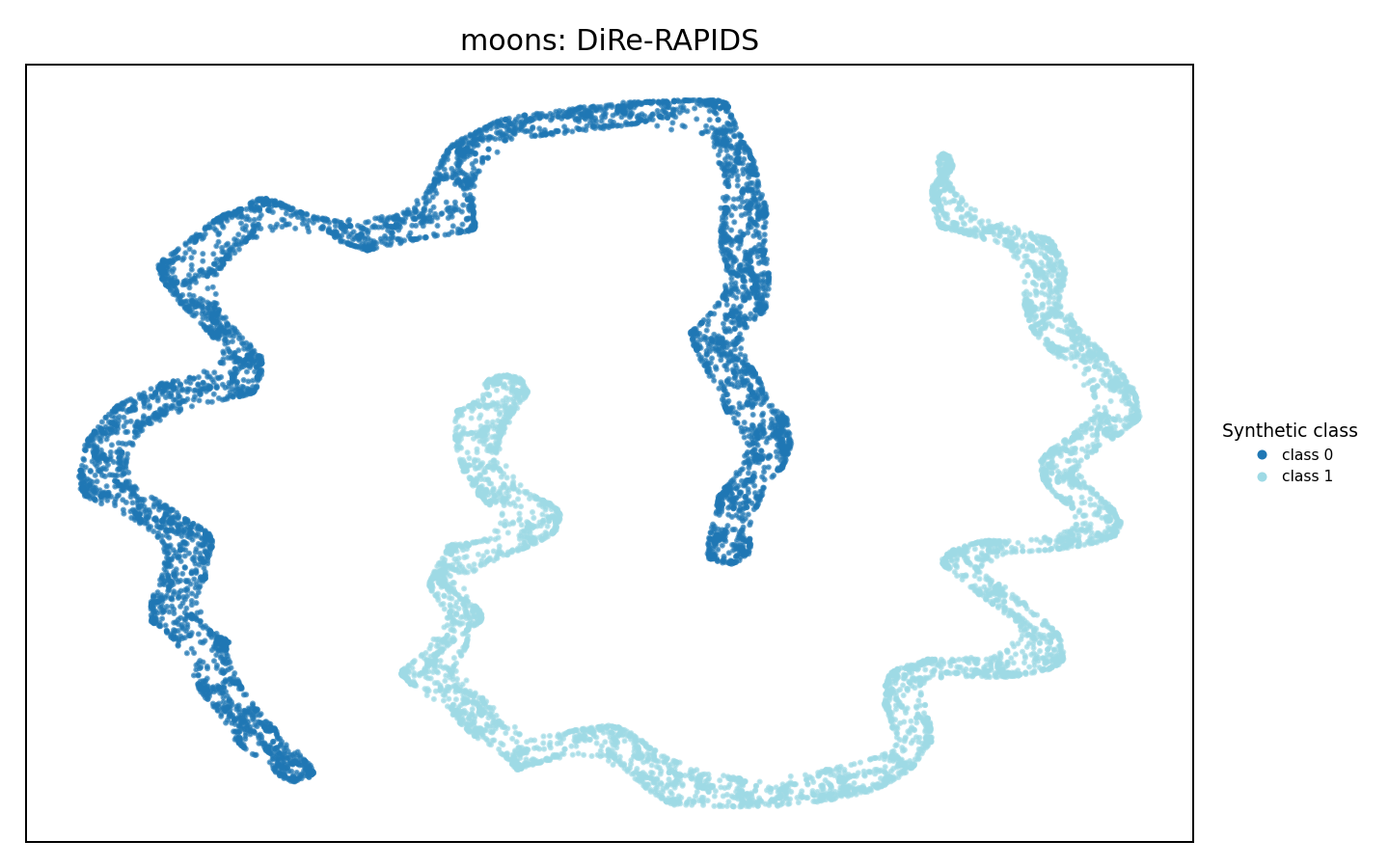}
        \caption{DiRe--RAPIDS}
        \label{fig:moons-dire-rapids}
    \end{subfigure}
    \hfill
    \begin{subfigure}[b]{0.45\textwidth}
        \centering
        \includegraphics[width=\textwidth]{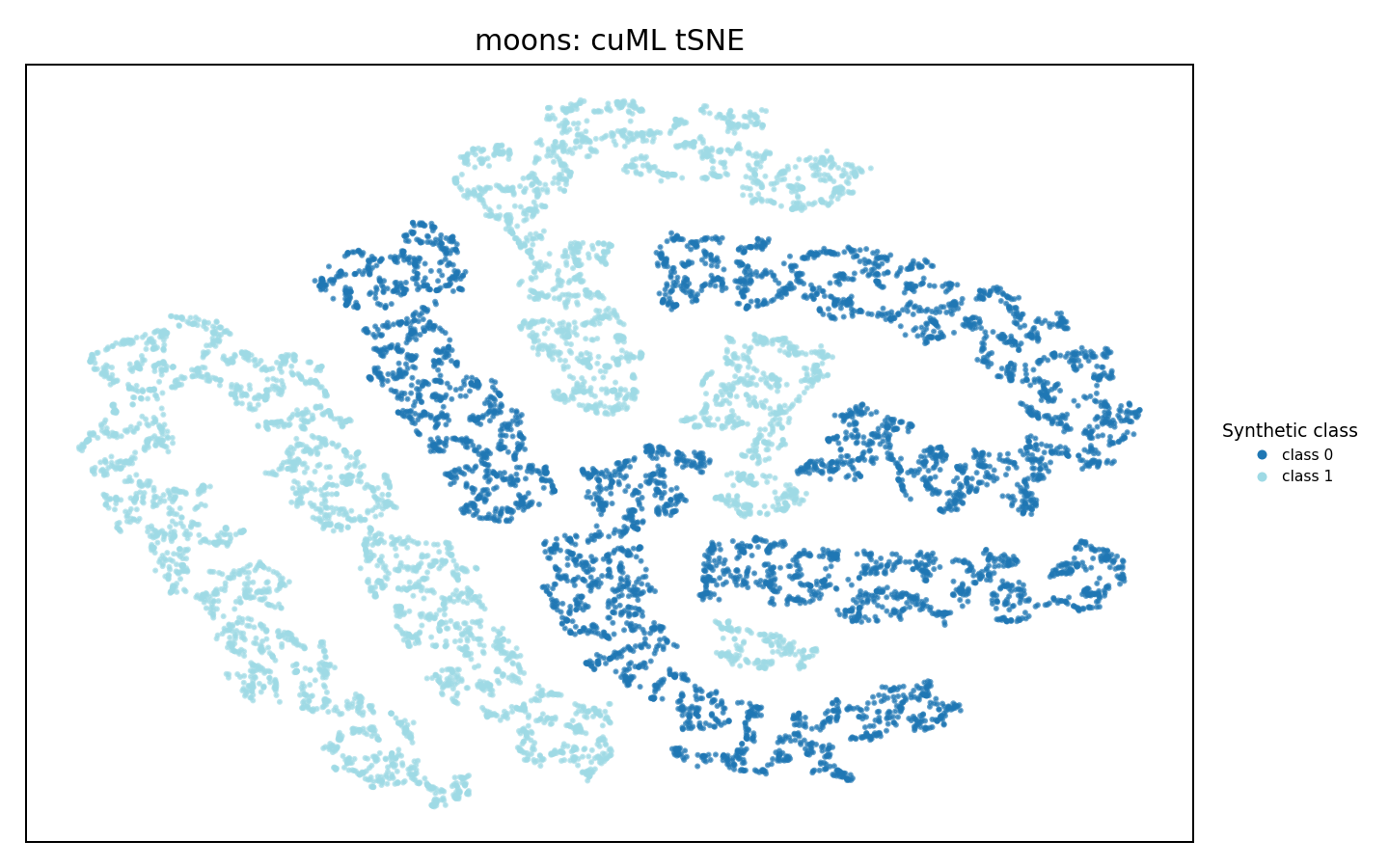}
        \caption{cuML tSNE}
        \label{fig:moons-tsne}
    \end{subfigure}
    \vspace{1em}
    \begin{subfigure}[b]{0.45\textwidth}
        \centering
        \includegraphics[width=\textwidth]{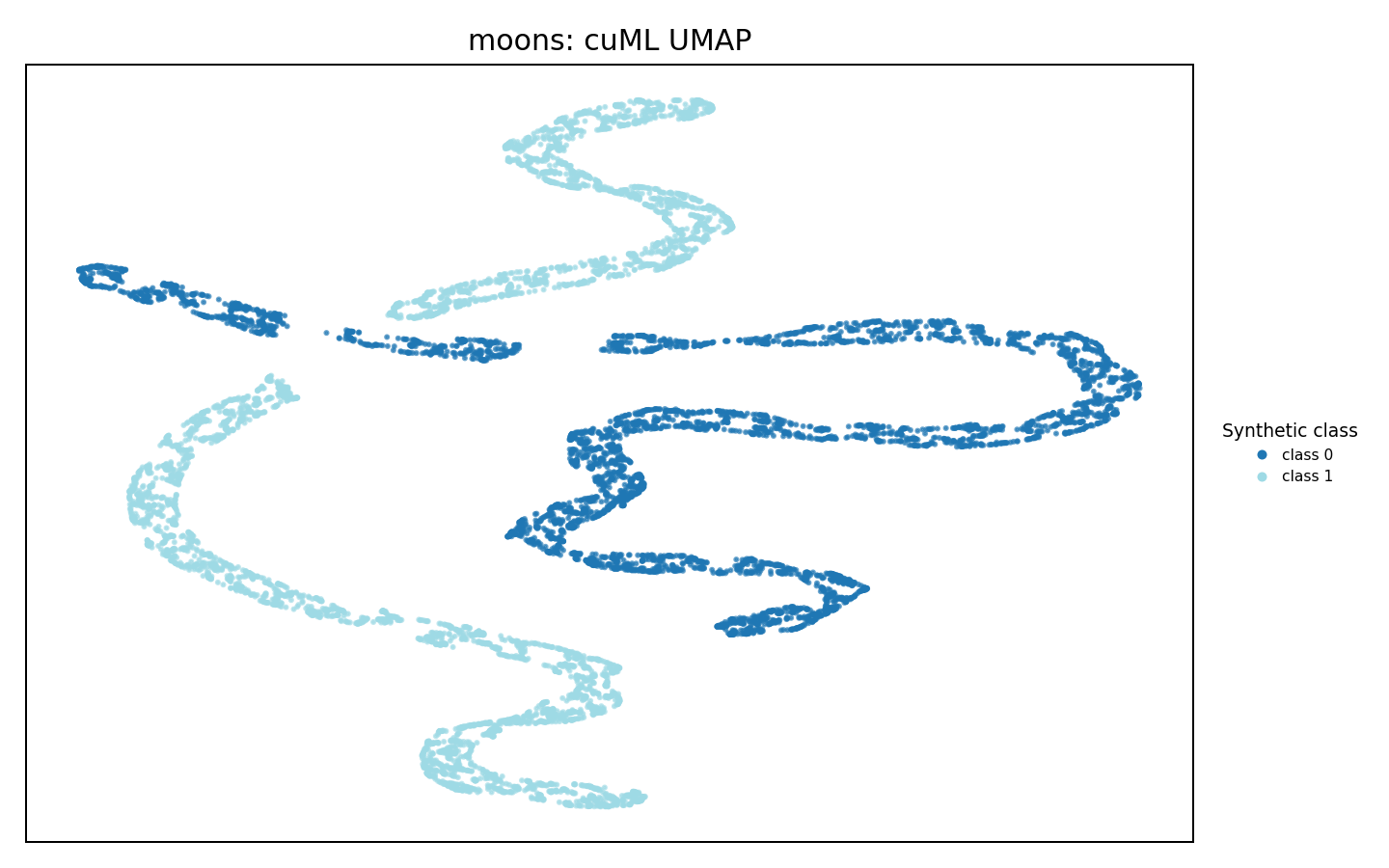}
        \caption{cuML UMAP}
        \label{fig:moons-cuml-umap}
    \end{subfigure}
    \hfill
    \begin{subfigure}[b]{0.45\textwidth}
        \centering
        \includegraphics[width=\textwidth]{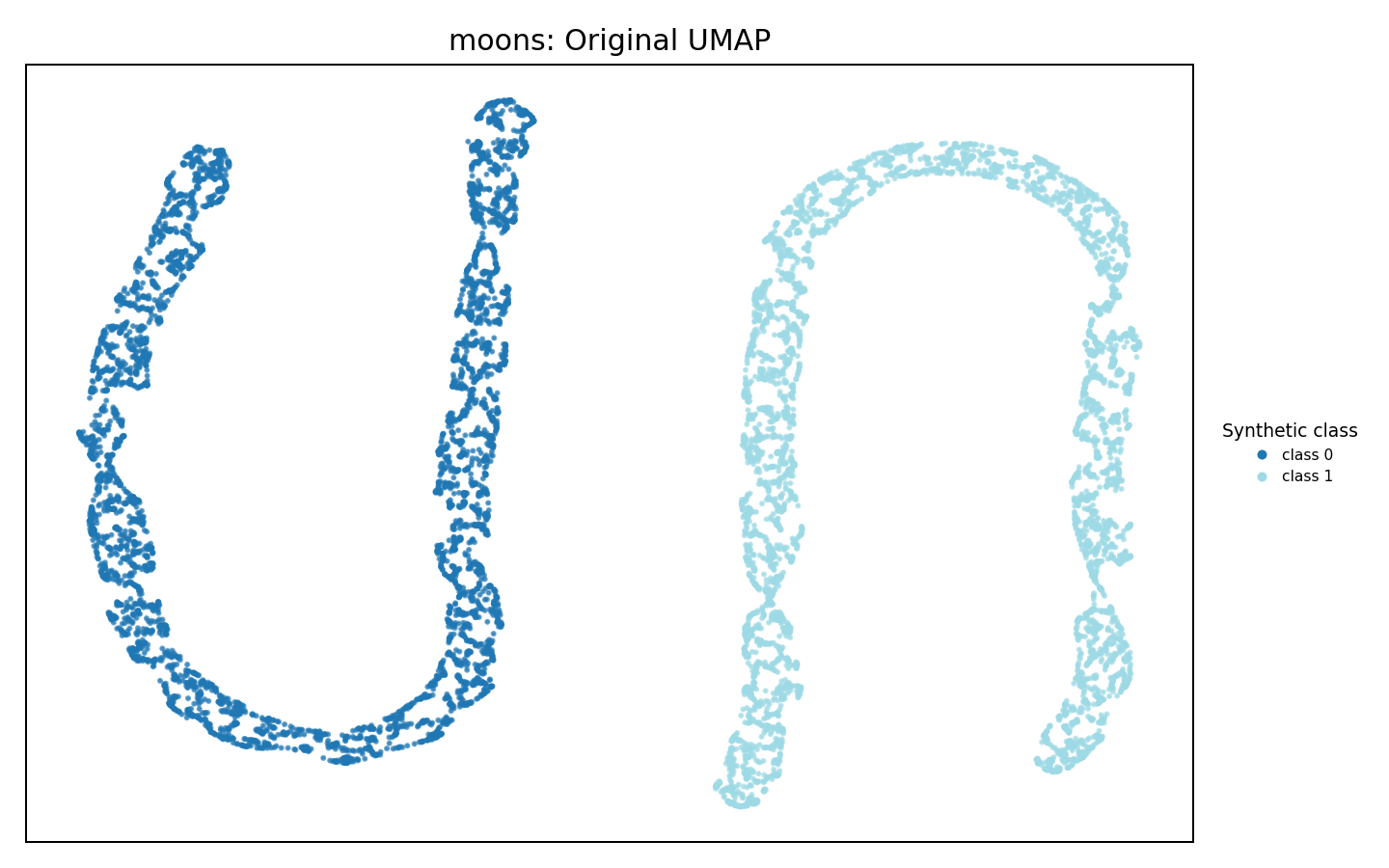}
        \caption{Original UMAP}
        \label{fig:moons-umap}
    \end{subfigure}
    \caption{"Two Half--Moons" dataset embeddings}
    \label{fig:moons}
\end{figure}

We compare the embeddings against both context metrics and persistence metrics, rather than relying on visual inspection alone.

\subsection{Dataset: Levine 13}

The Levine 13 CyTOF single--cell dataset serves as a biological benchmark in which preprocessing choices matter. In this benchmark, the expression channels are transformed with an inverse hyperbolic sine transform with cofactor 5; label and individual metadata columns are excluded from the feature matrix, and unassigned cells are removed.

\begin{figure}[ht]
    \centering
    \begin{subfigure}[b]{0.45\textwidth}
        \centering
        \includegraphics[width=\textwidth]{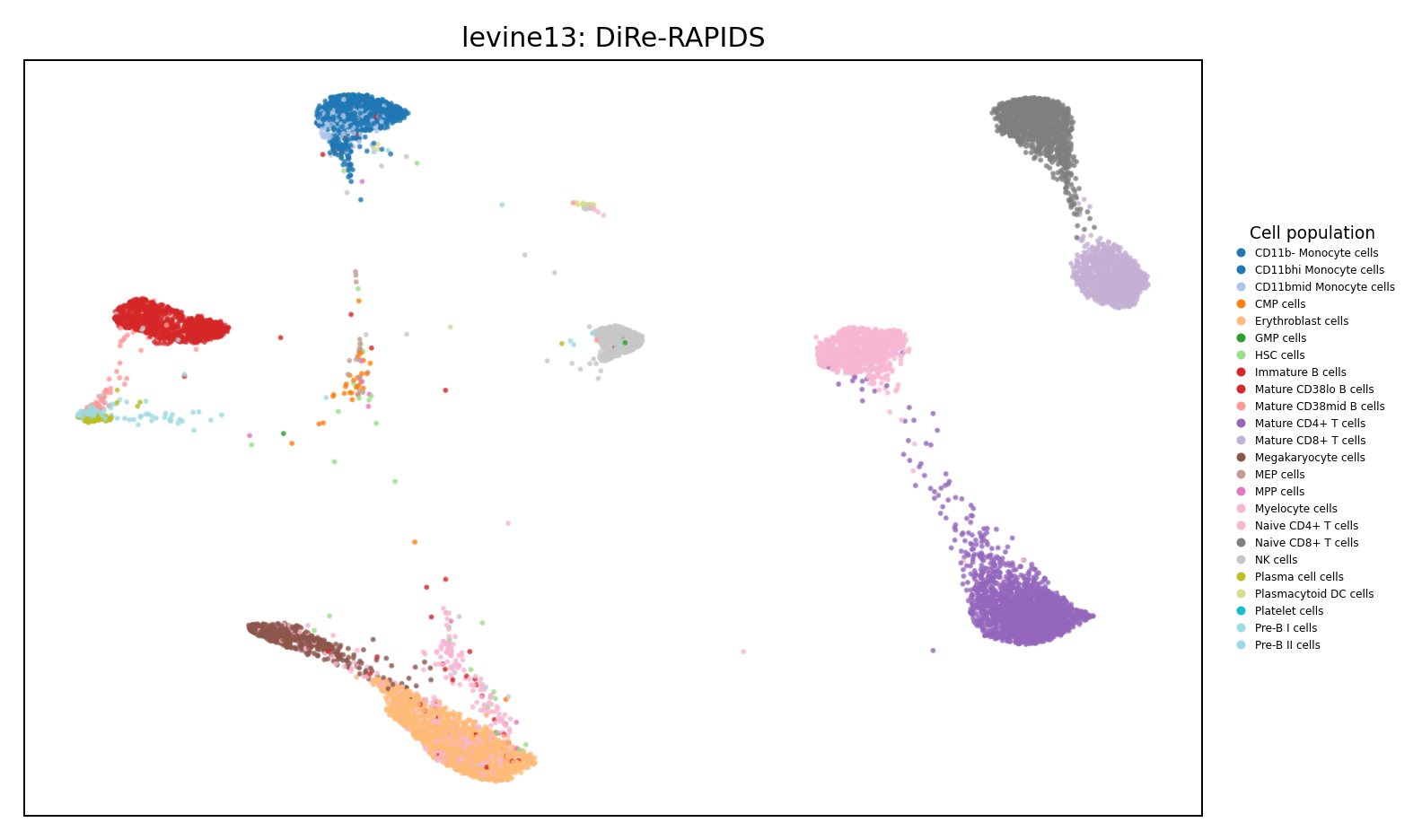}
        \caption{DiRe--RAPIDS}
        \label{fig:lev13-dire-rapids}
    \end{subfigure}
    \hfill
    \begin{subfigure}[b]{0.45\textwidth}
        \centering
        \includegraphics[width=\textwidth]{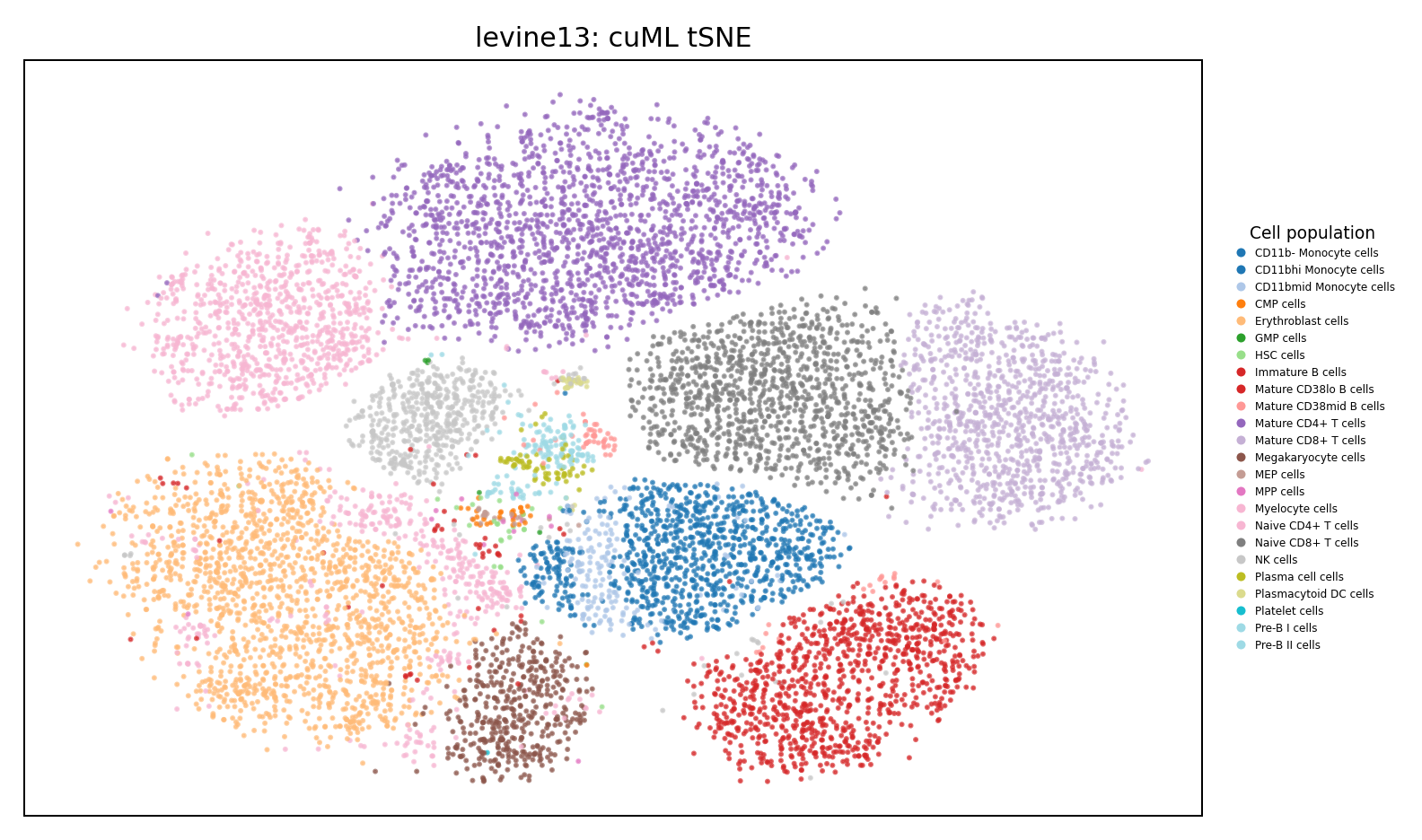}
        \caption{cuML tSNE}
        \label{fig:lev13-tsne}
    \end{subfigure}
    \vspace{1em}
    \begin{subfigure}[b]{0.45\textwidth}
        \centering
        \includegraphics[width=\textwidth]{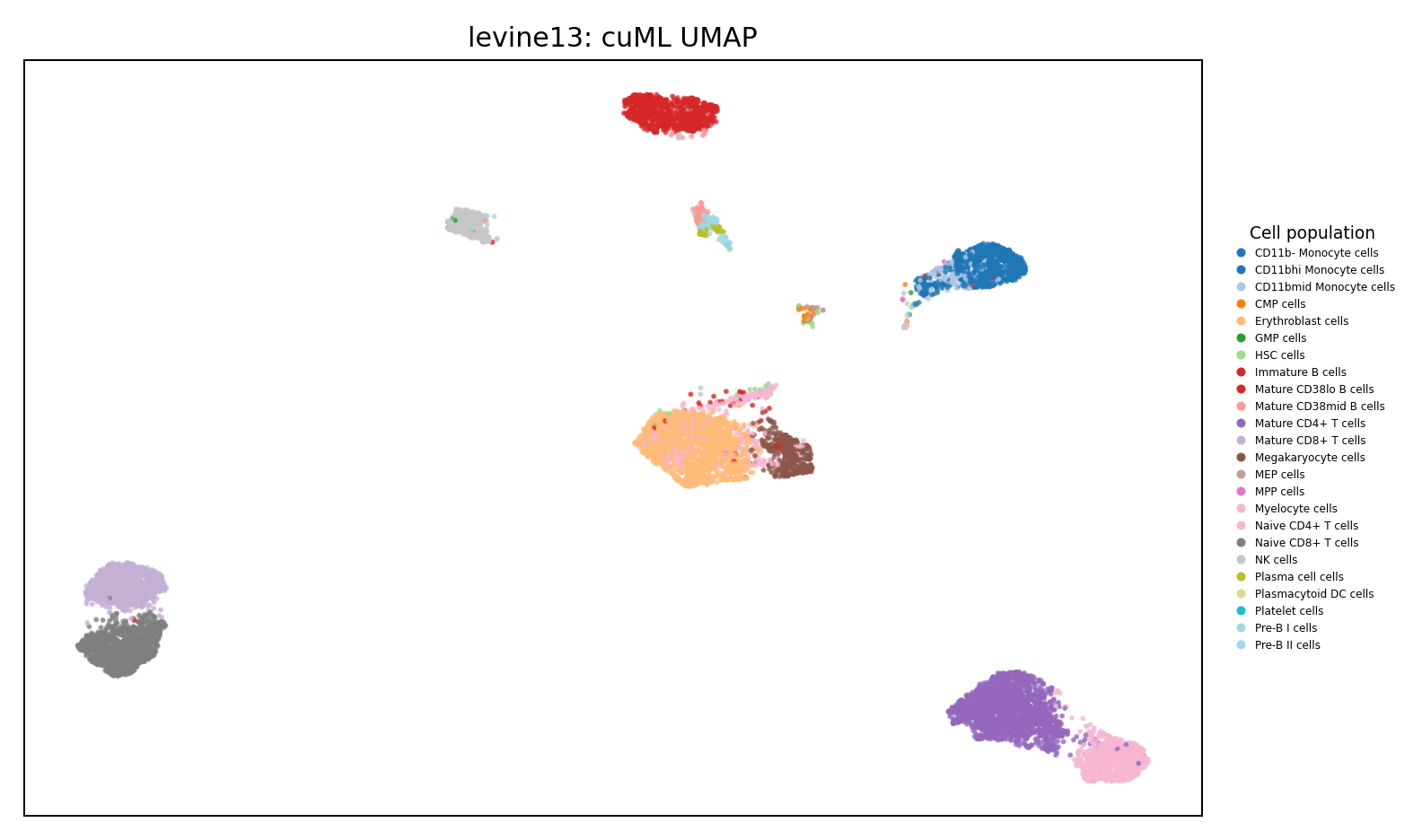}
        \caption{cuML UMAP}
        \label{fig:lev13-cuml-umap}
    \end{subfigure}
    \hfill
    \begin{subfigure}[b]{0.45\textwidth}
        \centering
        \includegraphics[width=\textwidth]{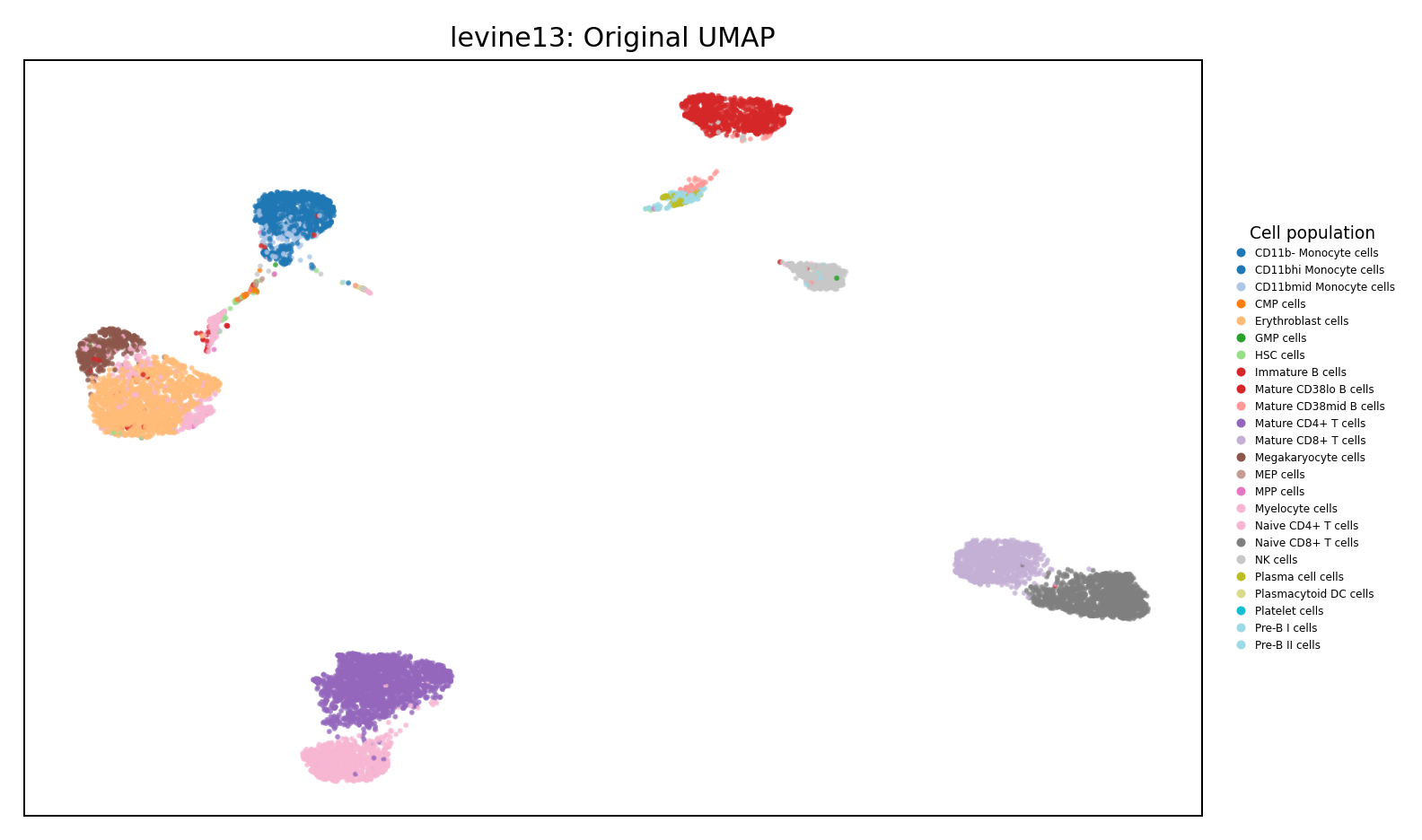}
        \caption{Original UMAP}
        \label{fig:lev13-umap}
    \end{subfigure}
    \caption{"Levine 13" dataset embeddings}
    \label{fig:levine13}
\end{figure}

We report persistence and context metrics separately, since biological labels, local neighborhoods, and topological summaries need not agree perfectly.

The generated per--metric Levine13 panels and raw repeated values are available
in the reproducibility package.

\subsection{Dataset: Levine 32}

The Levine 32 CyTOF dataset is handled with particular care because patient, batch, or other metadata columns can dominate an embedding if they are inadvertently included as numerical features. The same preprocessing protocol is used as for Levine 13: expression channels are arcsinh--transformed with cofactor 5, label and individual metadata columns are excluded from the feature matrix, and unassigned cells are removed.

\begin{figure}[ht]
    \centering
    \begin{subfigure}[b]{0.45\textwidth}
        \centering
        \includegraphics[width=\textwidth]{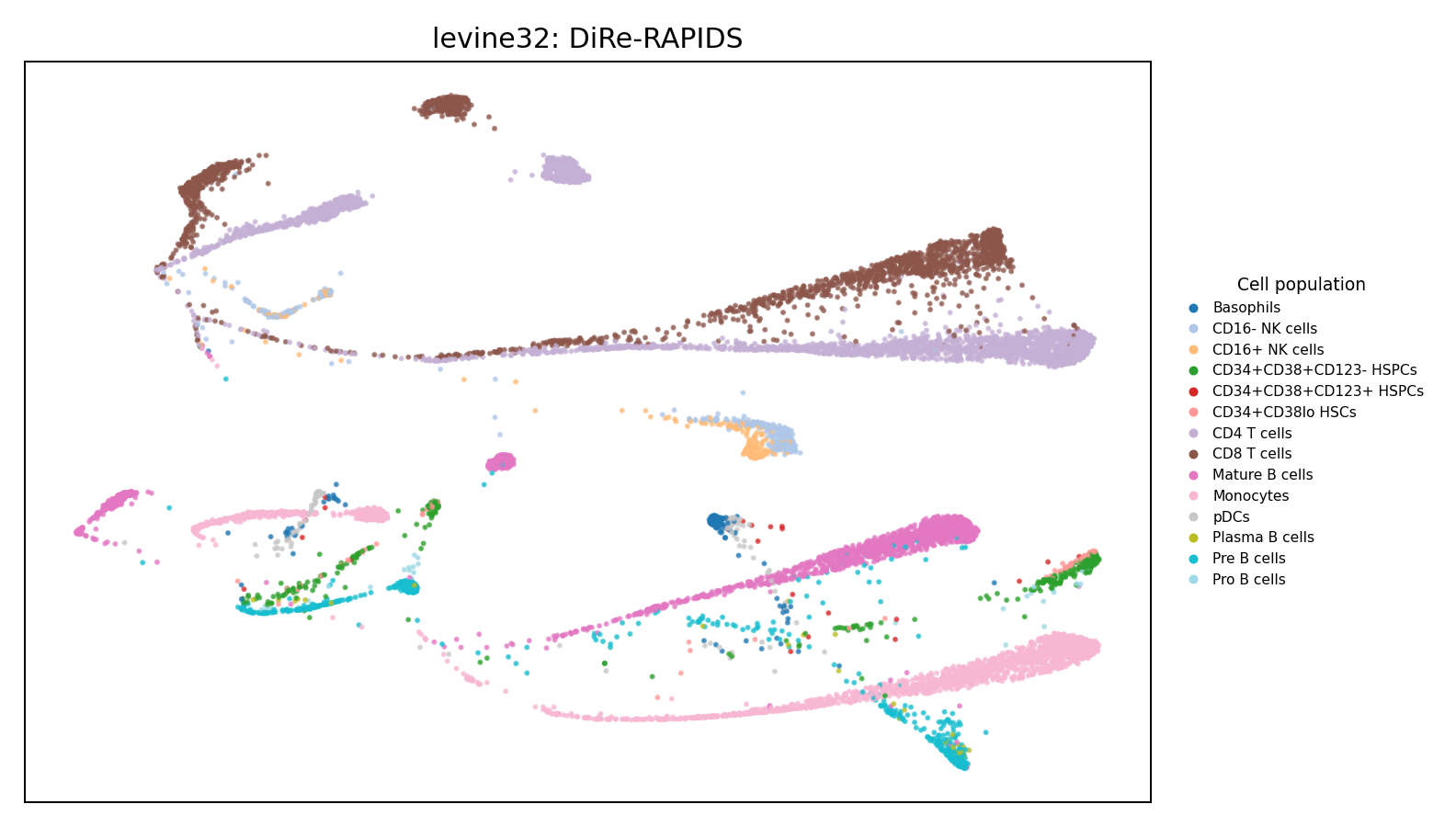}
        \caption{DiRe--RAPIDS}
        \label{fig:lev32-dire-rapids}
    \end{subfigure}
    \hfill
    \begin{subfigure}[b]{0.45\textwidth}
        \centering
        \includegraphics[width=\textwidth]{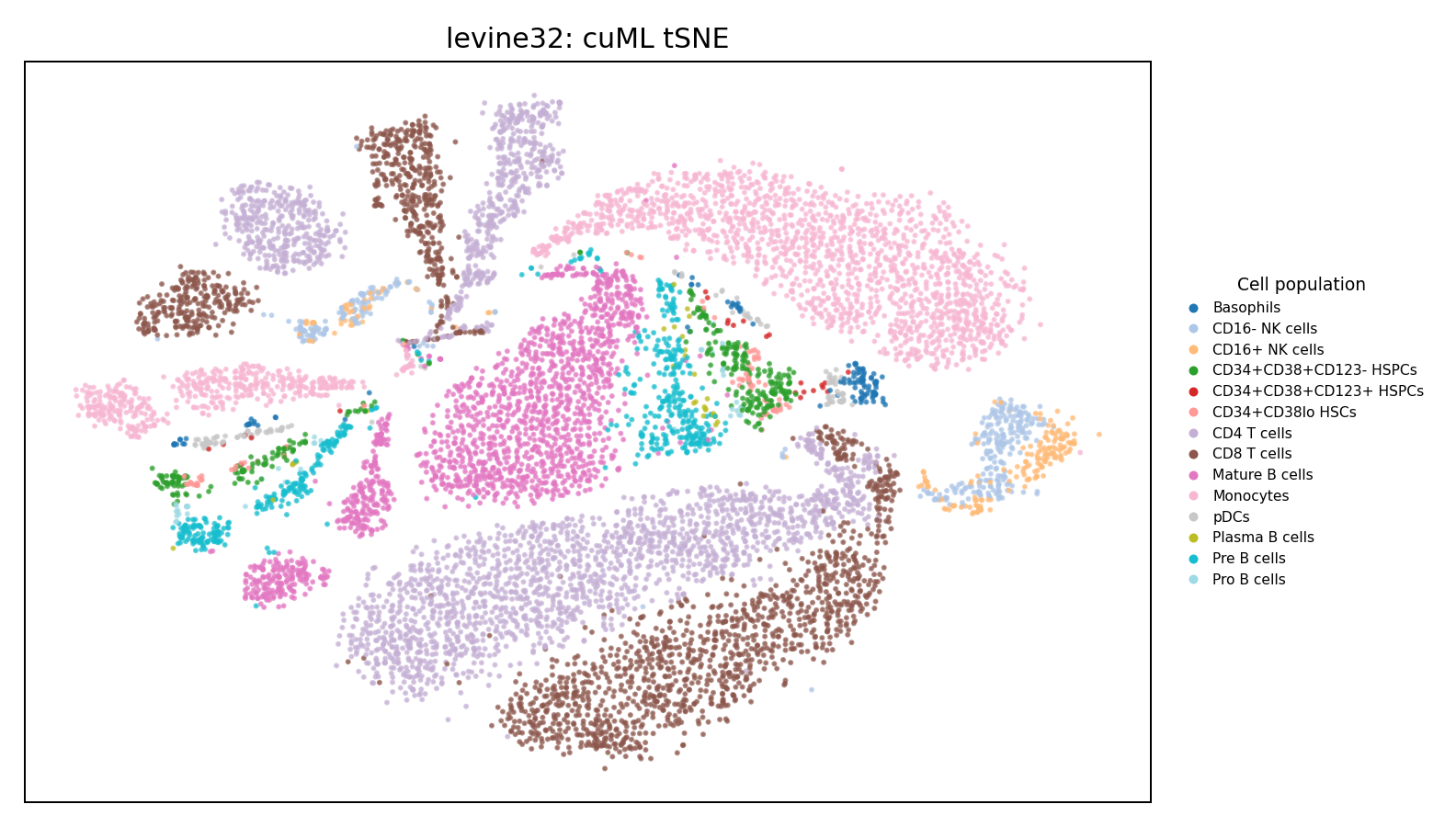}
        \caption{cuML tSNE}
        \label{fig:lev32-tsne}
    \end{subfigure}
    \vspace{1em}
    \begin{subfigure}[b]{0.45\textwidth}
        \centering
        \includegraphics[width=\textwidth]{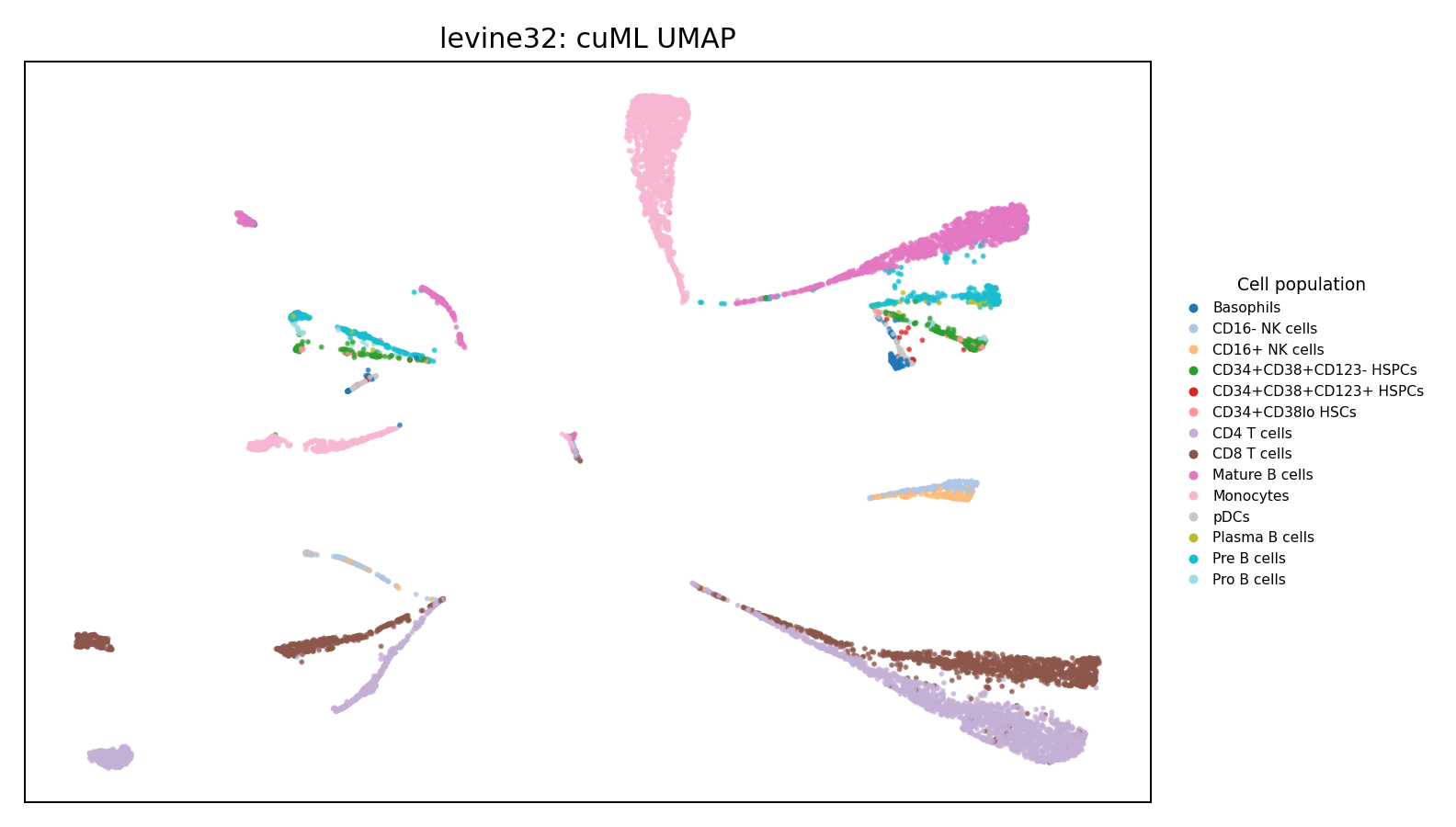}
        \caption{cuML UMAP}
        \label{fig:lev32-cuml-umap}
    \end{subfigure}
    \hfill
    \begin{subfigure}[b]{0.45\textwidth}
        \centering
        \includegraphics[width=\textwidth]{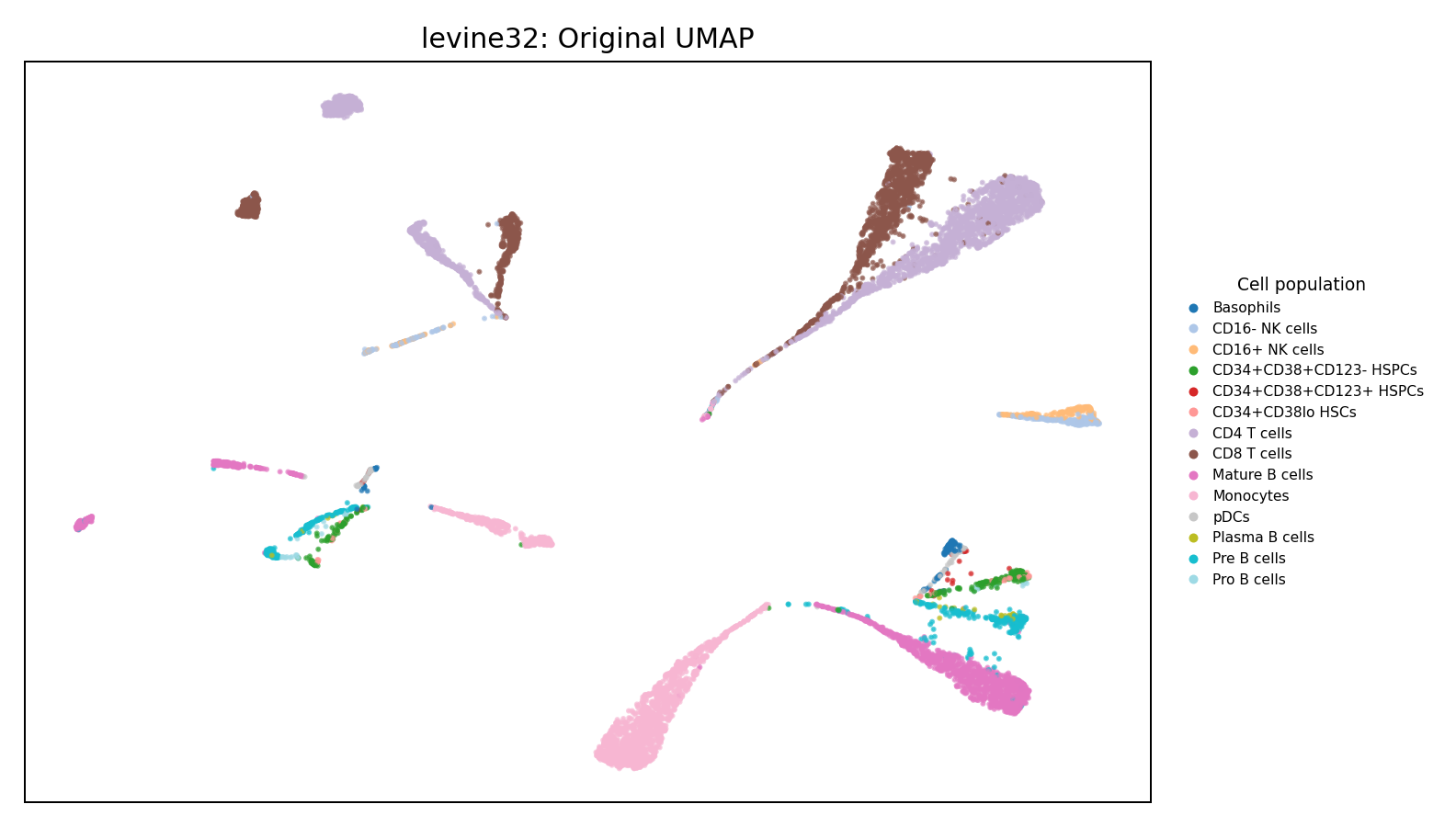}
        \caption{Original UMAP}
        \label{fig:lev32-umap}
    \end{subfigure}
    \caption{"Levine 32" dataset embeddings}
    \label{fig:levine32}
\end{figure}

The generated per--metric Levine32 panels and raw repeated values are available
in the reproducibility package.

\section{Scope and limitations}

DiRe retains the general framing stated in the abstract and introduction: it
is a dimensionality--reduction framework for a user--specified target
dimension $d$, including non--visual dimensions used in subsequent analysis.
The visual comparison figures in this study happen to instantiate $d=2$; that
experimental parameter neither defines the method nor restricts its intended
applications. As with any empirical study, the numerical rankings reported
here describe the recorded datasets, implementations, and configurations
rather than a universal ordering of all dimensionality--reduction methods.

No scalar metric used here defines ``global structure'' completely. Persistent
homology summarizes connectivity and loops across filtration scales but does
not retain every geometric relation; class--centroid correlations depend on
the supplied labels; context classification can reward separability that was
not present in the input; and neighborhood overlap is deliberately local. We
therefore report these diagnostics separately and do not infer universal
superiority from any one of them.

Persistent homology and exact neighbor comparisons are evaluated on fixed,
recorded samples for computational feasibility. These estimates have sampling
variation, addressed by repeated seeds, but they are not full--data persistence
computations. Biological cluster labels and marker statistics in the large
single--cell example come from the released Cell Ranger analysis. They support
comparison and interpretation of layouts, not a claim that a visualization
alone discovers a new cell type or establishes a causal biological mechanism.

Runtime results are implementation-- and hardware--specific. Same--GPU
comparisons control the accelerator for the released GPU implementations, while
CPU reference timings answer a different practical question and are displayed
separately. Approximate kNN search introduces a recall--speed tradeoff. The
large profile therefore reports both the production automatic cuVS policy and
a forced IVF--Flat control, their effective index types, stage times, and
fixed--query graph overlap; neither configuration is claimed optimal for every
dataset. Initialization, random seed, neighbor
count, force parameters, and iteration budget can all change a layout; the
initialization ablation measures part, but not all, of that sensitivity.

\section{Conclusions}

A new dimensionality reduction framework, called \texttt{DiRe}, has been developed with an explicit emphasis on preserving and measuring the dataset's homological structure. The latter is quantified by persistent homology and expressed via persistence diagrams or Betti curves. The benchmarks and metrics provided give evidence that DiRe can preserve global structure in cases where purely local criteria are insufficient, while also showing that no single metric captures every scientifically relevant aspect of an embedding.

The released default layout uses sparse $k$NN attraction and bounded
random--negative repulsion. This implementation distinction matters: the
layout is linear in the number of points for fixed protocol parameters and
does not reintroduce exact all--pairs repulsion. DiRe accepts a user--specified
target dimension $d$. The particular comparisons reported here use $d=2$ and
show dataset--dependent tradeoffs among local, labelled-context,
centroid-geometry, and persistent-homology diagnostics; that benchmark
configuration does not recast DiRe as a visualization-only method.

The DiRe software includes DiRe--RAPIDS \cite{dire-rapids} and DiRe--JAX \cite{dire-github,dire-joss}.

The benchmark logs, archived embedding images, plotting scripts, manuscript
build scripts, and Sphinx version of this manuscript are available in a
separate reproducibility repository \cite{homological-stability-repro}. The
large--data profile pins the DiRe commit and public dataset revisions, runs
method/size configurations in isolated processes, packages checksummed results,
and regenerates manuscript figures and tables locally. GPU acceleration is
required for reproducing the same--accelerator large--scale comparison; it is
not required to inspect the records or rebuild the manuscript artifacts.

\section{Funding} This work is supported by the Google~Cloud Research Award number GCP19980904. The computational benchmarks were run in part on Lambda Labs GPU infrastructure; we thank Lambda Labs for supporting this work with access to GPU computing resources.

\section{Author Contributions} A.K. and I.R. developed the necessary theoretical background and main algorithm. A.K. produced and prepared figures. All authors reviewed the manuscript.

\newpage

\begingroup
\sloppy
\printbibliography
\endgroup

\newpage

\appendix
\section{Released 10x marker annotation}

The following marker statistics are copied from the pinned public Cell Ranger
analysis and are included to make the cluster-level visual and adjacency audit
interpretable. They were not inferred from any of the compared layouts.

\begin{longtable}{@{}rrp{0.62\linewidth}@{}}
\caption{Released 10x cluster sizes and the five strongest positive differential-expression markers used to annotate the million-cell visual audit.}\label{tab:revision3-tenx-markers}\\
\toprule
Official cluster & Cells & Top released positive markers \\
\midrule
\endhead
1 & 331,960 & Tuba1a, Gpm6a, Cd24a, Gria2, Tsc22d1 \\
2 & 196,425 & Igfbpl1, Sox11, Pcp4, Nfib, H3f3b \\
3 & 118,140 & Snca, Zbtb20, Crym, Stmn2, Nts \\
4 & 110,356 & Gria2, Ptprd, Arpp21, Tmem108, Nts \\
5 & 105,158 & Maf, Nxph1, Meg3, Sst, Npy \\
6 & 88,009 & Neurog2, Eomes, Gadd45g, Mdk, Mfap4 \\
7 & 84,088 & Nnat, Dlx6os1, Sp9, Gad2, Ccnd2 \\
8 & 80,672 & Fabp7, Dbi, Mt3, Aldoc, Vim \\
9 & 51,662 & 2810417H13Rik, H2afz, Rrm2, Hmgb2, Ccnd2 \\
10 & 49,421 & Ube2c, Hmgb2, Cks2, Cenpf, Ccnb1 \\
11 & 20,993 & Snhg11, Reln, Ndnf, Meg3, Pcp4 \\
12 & 19,814 & Hbb-bs, Hba-a1, Hba-a2, Hbb-bt, Alas2 \\
13 & 15,269 & Cldn5, Igfbp7, Bsg, Itm2a, Flt1 \\
14 & 11,697 & Olig1, Pdgfra, Serpine2, Olig2, Lhfpl3 \\
15 & 11,137 & Rgs5, Higd1b, Vtn, Igf2, Ndufa4l2 \\
16 & 5,525 & Ccl4, C1qb, C1qc, C1qa, Ccl3 \\
17 & 5,038 & Ttr, Dbi, Rspo1, Wnt8b, Apoe \\
18 & 589 & Meig1, Dynlrb2, Foxj1, Mt2, Mt1 \\
19 & 123 & Plac8, Lyz2, S100a8, Ifi27l2a, Cd52 \\
20 & 51 & S100a9, S100a8, BC100530, Ngp, Stfa1 \\
\bottomrule
\end{longtable}

\newpage

\end{document}